\documentclass[twoside,11pt]{article}

\usepackage{jmlr2e}
\usepackage[utf8]{inputenc} 
\usepackage[T1]{fontenc}    
\usepackage{hyperref}
\usepackage{url}            
\usepackage{booktabs}       
\usepackage{amsfonts}       
\usepackage{nicefrac}       
\usepackage{tikz} 
\usepackage{subcaption}
\usepackage{microtype}      
\usepackage{xcolor}         
\usepackage{relsize}
\usepackage{enumitem}
\usepackage{mathtools} 
\usepackage{amssymb}
\usepackage{cleveref}
\usepackage{bbm}

\newcommand{\lambdaijr}{\lambda_{ij}^{\mathsmaller{(1)}}}
\newcommand{\lambdaijg}{\lambda_{ij}^{\mathsmaller{(2)}}}

\newcommand{\gammaijr}{\gamma^{\mathsmaller{(1)}}_{ij}}
\newcommand{\gammaijg}{\gamma^{\mathsmaller{(2)}}_{ij}}

\newcommand{\yijmm}{y^{\mathsmaller{(1)}}_{ij}}
\newcommand{\yijuu}{y^{\mathsmaller{(2)}}_{ij}}
\newcommand{\yijm}{y^{(\textrm{1})}_{ij}}
\newcommand{\yiju}{y^{(\textrm{2})}_{ij}}

\newcommand{\yicjkmm}{y^{\mathsmaller{(1)}}_{icjk}}

\newcommand{\tminus}{\!-\!}
\newcommand{\tplus}{\!+\!}
\newcommand{\compcond}[1]{\left(#1 \given-\right)}

\newcommand{\given}{\,|\,}

\newcommand{\teq}{\!=\!}

\newcommand{\thetaic}{\theta_{ic}}

\newcommand{\pick}[1]{\pi_{ck}^{\mathsmaller{(#1)}}}

\newcommand{\eps}[1]{\epsilon_0^{\mathsmaller{(#1)}}}
\newcommand{\lamij}[1]{\lambda_{ij}^{\mathsmaller{(#1)}}}

\newcommand{\alphaij}[1]{\alpha_{ij}^{\mathsmaller{(#1)}}}

\newcommand{\Pois}[1]{\textrm{Pois}\left( #1 \right)}

\newcommand{\Gam}[1]{\textrm{Gam}\left( #1 \right)}

\newcommand{\DNCB}[1]{\textrm{DNCB}\left( #1 \right)}

\jmlrheading{}{}{}{}{}{Anjali N. Albert, Patrick Flaherty, and Aaron Schein}

\firstpageno{1}
\ShortHeadings{Stable Dimensionality Reduction for Bounded Support Data}{Albert et. al.}

\begin{document}

\title{Doubly Non-Central Beta Matrix Factorization for Stable Dimensionality Reduction of Bounded Support Matrix Data}

\author{\name Anjali N. Albert \email gnagulpally@umass.edu \\
       \addr Department of Mathematics \& Statistics\\
       University of Massachusetts\\
       Amherst, MA 01003, USA
       \AND
       \name Patrick Flaherty \email pflaherty@umass.edu \\
       \addr Department of Mathematics \& Statistics\\
       University of Massachusetts\\
       Amherst, MA 01003, USA
       \AND
       \name Aaron Schein \email schein@uchicago.edu \\
       \addr Department of Statistics \& Data Science Institute\\
       University of Chicago\\
       Chicago, IL 60637, USA
}

\editor{}

\maketitle

\begin{abstract}
We consider the problem of developing interpretable and computationally efficient matrix decomposition methods for matrices whose entries have bounded support.
Such matrices are found in large-scale DNA methylation studies and many other settings.
Our approach decomposes the data matrix into a Tucker representation wherein the number of columns in the constituent factor matrices is not constrained.
We derive a computationally efficient sampling algorithm to solve for the Tucker decomposition.
We evaluate the performance of our method using three criteria: predictability, computability, and stability.
Empirical results show that our method has similar performance as other state-of-the-art approaches in terms of held-out prediction and computational complexity, but has significantly better performance in terms of stability to changes in hyper-parameters.
The improved stability results in higher confidence in the results in applications where the constituent factors are used to generate and test scientific hypotheses such as DNA methylation analysis of cancer samples.
\end{abstract}


\section{Introduction}

Bounded-value data is common in many data science applications.
In computational biology, DNA methylation data is $(0,1)$-bounded because each entry is a measurement of the fraction of CpG sites that are methylated at a genetic locus in a sample~\citep{shafi2018survey}.
Collaborative filtering and recommendation systems make use of data on pairs of entities (e.g. users and movies) whose value is an integer from one to five~\citep{bennett2007netflix}. 
In these fields, a central scientific goal is to find a low-dimensional representation of the high-dimensional data that yields meaningful and reproducible scientific insights.

 Matrix decomposition methods are widely used to learn low-dimensional representations~\citep{lee1999learning, roweis2000nonlinear, van2008visualizing}.
Tensor decomposition methods extend matrix decomposition methods to multi-dimensional data and have been used in computer vision, graph analysis, signal processing and many other applications~\citep{kolda2009tensor}.
It was originally developed as a method for extending factor analysis to three-mode tensors~\citep{tucker1966some}.
It was later extended to generalize singular value decomposition and principal component analysis to higher-order tensors~\citep{grasedyck2010hierarchical}.
Recently, Tucker decomposition has been combined with ideas in sketching~\citep{malik2018low} and randomized linear algebra to improve computational efficiency for large-scale data sets~\citep{minster2020randomized}.

Following the notation in \citet{minster2020randomized}, the Tucker decomposition represents a $d$-mode tensor $\boldsymbol{\mathcal{X}} \in \mathbb{R}^{I_1 \times \cdots \times I_d}$ as the tensor product between a lower-dimensional core matrix $\mathcal{G} \in \mathbb{R}^{r_1 \times \cdots \times r_d}$ and factor matrices $\{ \mathbf{A}_j\}_{j=1}^d$ where $\mathbf{A}_j \in \mathbb{R}^{I_j \times r_j}$.
The decomposition is such that $\mathcal{X} = \mathcal{G} \bigtimes_{j=1}^d \mathbf{A}_j$ and the Tucker representation of $\mathcal{X}$ is written compactly as $[\mathcal{G}; \mathbf{A}_1, \ldots, \mathbf{A}_d]$.

Other tensor representations exist: CANDECOMP/PARAFAC  (CP)~\citep{kolda2009tensor}, hierarchical  Tucker~\citep{grasedyck2010hierarchical},  and  tensor train~\citep{oseledets2011tensor}.   
The  CP representation  decomposes the tensor into an outer product of rank-1 tensors.
For 2-mode tensors typically found in recommendation problems and topic models, Tucker decomposition has some unique benefits compared to other alternatives.
Matrix factorization methods that represent a data table as product of two matrices constraint the number of columns in the first factor matrix to be the same as the number of rows in the second factor matrices.
For example, in a topic model decomposition of a document-term matrix, the number of groups of documents is constrained to be equal to the number of topics.
This constraint is often not realistic in practice and undesirable when interpreting the model.
Tucker decomposition has the benefit, due to the core matrix, that the reduced ranks of the factor matrices are decoupled from one another making for more interpretable models. 

Singular value decomposition of real-valued matrices to a Tucker representation can be accomplished using higher order SVD (HOSVD) and sequentially  truncated higher order SVD (STHOSVD)~\citep{minster2020randomized}.
Recently, Nonnegative Tucker decomposition (NTD) has been developed for non-negative-valued matrices and tensors~\citep{kim2007nonnegative}.
NTD has proven to be empirically useful in identifying meaningful latent features in image data~\citep{zhou2015efficient}.
Using the fact that all entries are nonnegative leads to dramatic improvements in computational efficiency~\citep{wang2012nonnegative}.
However, there are not yet any methods, to our knowledge, for finding a Tucker decomposition of tensors whose entries are $(0,1)$-bounded.

Matrix factorization and Tucker decomposition is particularly relevant for DNA methylation data sets.
DNA methylation enables the cell to exert epigenetic control over the expression of genetic material to affect cellular processes such as development, inflammation, and genomic imprinting~\citep{bock2012analysing}.
There is evidence that has associated DNA methylation patterns with environmental exposures, and aberrant DNA methylation patterns have been used as biomarkers for cancer prognosis~\citep{laird2003power}.
DNA methylation data can be represented as an $I \times J$ matrix where an entry, $x_{ij}$ is the proportion (bounded by $(0,1)$) of CpG sites that are methylated in sample $i$ at CpG island or ``locus'' $j$.

A particular application of the analysis of bounded support DNA methylation data is in continuous monitoring of cell-free DNA for metastatic cancer recurrence~\citep{hidalgo2022cellfree}.
A primary tumor may be treated through surgical or other means into remission, but the possibility of a distant metastatic recurrence remains.
DNA methylation patterns are highly correlated with cell-type and thus tissue of origin of the cell~\citep{kaplan2023methylation}.
Therefore, DNA methylation patterns preserved in circulating cell-free DNA represent an opportunity to monitor and possibly localize metastatic recurrence in a non-invasive manner~\citep{ji2023cellfree}.
Several recent papers have developed matrix factorization for DNA methylation data.
\citet{yoo2009probabilistic} develop a matrix tri-factorization model without priors and estimate the factor matrices with an EM algorithm.
\citet{vcopar2017scalable} is concerned primarily with scalability and develop a block-wise update algorithm that can be parallelized on a GPU.
Finally, \citet{park2019bayesian} use an exponential prior on the elements of the middle factor in the tri-factorization, and a Gaussian prior for the right matrix. A variational inference algorithm is used to provide estimates of the latent matrices. 
The need for interpretable and predictive methods for analysis of this type of data motivates the need for computationally efficient methods for finding Tucker representations of bounded support data.

\paragraph{Contributions}

This paper significantly extends the work of \citet{schein2021doubly} to develop a family of hierarchical statistical models for matrix decomposition of bounded support data. These models are based on the doubly non-central beta (DNCB) distribution as a flexible likelihood term. We build this family of generative models to encompass both standard matrix factorization (DNCB-MF) and Tucker decomposition (DNCB-TD). We derive a fast Markov-chain Monte Carlo inference method for both matrix factorization and Tucker decomposition using an augment-and-marginalize approach. We measure prediction accuracy, computational efficiency, and stability to changes in hyper-parameters for DNCB-MF and DNCB-TD. An analysis of real methylation data (both array-based and sequence-based) shows that DNCB-TD identifies coherent clusters of samples and clusters of features (pathways) that correlate with cancer development mechanisms. An analysis of facial image data shows that the model is flexible and can identify coherent clusters for a wide range of bounded support data distributions. Finally, we provide a replicable implementation at \url{https://github.com/flahertylab/dncb-matrix-fac}.

\section{Doubly Non-Central Beta Factorization}
\vspace{-0.5em}
\label{sec:model}
To analyze matrices of bounded support data, we introduce a family of probabilistic matrix decomposition models that are unified under the assumption that entries $\beta_{ij} \in (0,1)$ of an observed  $I \times J$ matrix are doubly non-central beta (DNCB) random variables,
\begin{align}
\label{eq:marglikelihood}
    \beta_{ij} &\sim \DNCB{\eps{1}, \eps{2}, \lamij{1}, \lamij{2}},
\end{align}
where the \textit{shape} parameters, $\eps{1}$ and $\eps{2}$, are shared across all $(i,j) \in [I] \times [J]$, and the \textit{non-centrality} parameters, $\lamij{1}$ and $\lamij{2}$, are unique to each $(i,j)$ and decompose linearly into latent factors---e.g., $\lambdaijr = \sum_{k=1}^{K}  \theta_{ik}^{\mathsmaller{(1)}}\phi_{kj}$. 
We introduce, later, two different ways of decomposing the non-centrality parameters (CP and Tucker decomposition) that yield qualitatively different interpretations of the latent factors.

\begin{figure}[ht]
\centering
\includegraphics[width=0.48\linewidth]{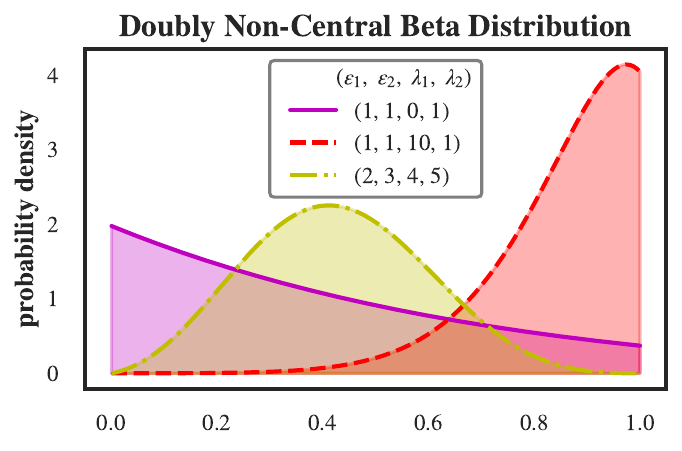} \hfill
\includegraphics[width=0.49\linewidth]{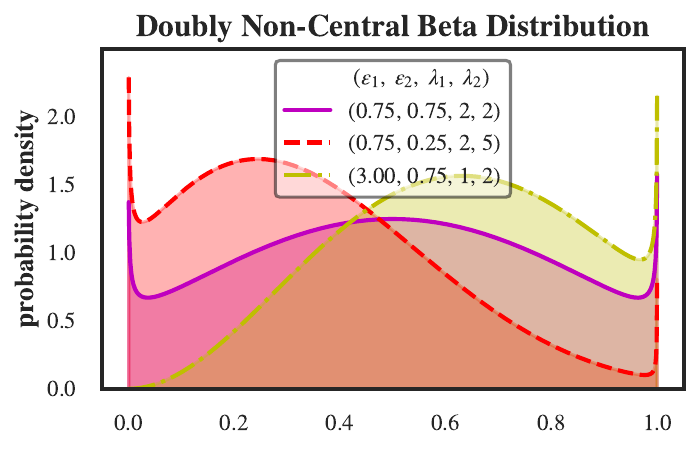} \hfill
\caption{The doubly non-central beta (DNCB) distribution can assume a shape like the standard beta (left). Alternatively, the DNCB distribution can take a multi-modal shape if $\epsilon_1 < 1$ or $\epsilon_2 < 1$ (right). This expressiveness is particularly useful when modeling DNA methylation datasets, which are typically highly dispersed and multi-modal. 
}
\label{fig:dncb}
\end{figure}

The DNCB distribution~\citep{johnson1995continuous} is a generalization of the standard beta distribution whose properties have been recently explicated by~\citet{orsi2015} and \citet{orsi2017new}. We provide a definition of the DNCB distribution here and visualize it in~\Cref{fig:dncb}.

\begin{definition}[Doubly non-central beta distribution] 
\label{df:dncb}
A doubly non-central beta random variable $\beta \sim \DNCB{\epsilon_1, \epsilon_2, \lambda_1, \lambda_2}$ is continuous with bounded support $\beta \in (0,1)$. Its distribution is defined by positive \emph{shape parameters} $\epsilon_1 > 0$ and $\epsilon_2 > 0$, non-negative \emph{non-centrality parameters} $\lambda_1 \geq 0$ and $\lambda_2 \geq 0$, and probability density function (PDF) equal to~\looseness=-1
\begin{equation}
\nonumber \DNCB{\beta;\,\epsilon_1, \epsilon_2, \lambda_1, \lambda_2} = \textrm{Beta}(\beta;\,\epsilon_1, \epsilon_2) \, e^{-\lambda_\bullet}\, \Psi_2\left[\epsilon_\bullet;\, \epsilon_1,\, \epsilon_2;\, \lambda_1\beta,\, \lambda_2(1-\beta)\right]
\end{equation}
where $\epsilon_\bullet \triangleq \epsilon_1 + \epsilon_2$, $\lambda_\bullet \triangleq \lambda_1 + \lambda_2$, and $\Psi_2\left[\cdot;\, \cdot,\, \cdot;\, \cdot,\, \cdot\right]$ is Humbert's confluent hypergeometric function.~\looseness=-1
\end{definition}

The DNCB distribution has several properties that make it useful for modeling bounded matrix data. 
First, as shown by by~\citet{orsi2015}, the density is well-behaved (has finite and positive density) when approaching the bounds of support (0 or 1), unlike the standard beta whose density goes to 0 or $+\infty$. This fact suggests that probabilistic models based on a DNCB likelihood may be more flexible and stable than ones based on the standard beta, particularly if data are highly dispersed around the extremes.
Second, the DNCB distribution yields several augmentations that are useful for building hierarchical Bayesian models and deriving efficient posterior inference algorithms, as we will show. 
In particular, the DNCB distribution has a mixture representation in terms of Poisson-distributed auxiliary variables. 
The DNCB random variable $\beta_{ij}$ given in \Cref{eq:marglikelihood} can be generated from a standard beta distribution, conditional on draws of two Poisson random variables:~\looseness=-1
\begin{equation}
\label{eq:betalikelihood}
\begin{aligned}
\beta_{ij} \mid \yijmm, \yijuu &\sim \textrm{Beta} \left(\eps{1} + \yijmm, \eps{2} + \yijuu\right), \\ 
y_{ij}^{\mathsmaller{(t)}} &\sim \Pois{\lambda_{ij}^{\mathsmaller{(t)}}} \quad t \in \{1,2\}.
\end{aligned}
\end{equation}

There is a large literature on latent factor and hierarchical Bayesian models based on the Poisson likelihood~\citep{cemgil09bayesian, gopalan2015poisson, zhounegative}. 
The Poisson enjoys closed-form conjugate priors, unlike the standard beta, alongside other well-understood properties and relationships that makes it easy to build flexible and tractable models. 
The mixture representation in \Cref{eq:betalikelihood} allows us to build tractable latent factor models for bounded support data by linking observed data $\beta_{ij}$ to latent count variables, $\yijm$ and $\yiju$, and then hierarchically modeling those counts using techniques and motifs from the literature on Poisson-based models.


Representing the DNCB distribution as a Poisson-Beta mixture allows for yet another useful augmentation based on the relationship between the Beta and Gamma distributions.
If $\gamma_1 \sim \Gam{a_1,\,c}$ and $\gamma_2 \sim \Gam{a_2,\,c}$ are independent Gamma random variables with possibly different shape parameters, $a_1$ and $a_2$, but shared rate parameter $c$, then the proportion $\beta = \nicefrac{\gamma_1}{\left( \gamma_1 + \gamma_2 \right)}$ is marginally Beta-distributed as $\beta \sim \textrm{Beta}(a_1,\,a_2)$.
A fully augmented version of the marginal likelihood in~\Cref{eq:marglikelihood} can therefore be written as 
\begin{equation}
\label{eq:augmentedlikelihood}
\begin{aligned}
    \beta_{ij} \mid \gammaijr, \gammaijg  &= \frac{\gammaijr} {\gammaijr + \gammaijg}  &\\
    \gamma_{ij}^{\mathsmaller{(t)}} \mid y_{ij}^{\mathsmaller{(t)}} &\sim \Gam{\eps{t} + y_{ij}^{\mathsmaller{(t)}}, c_j}  &\quad t &\in \{1,2\} \\
        y_{ij}^{\mathsmaller{(t)}} &\sim \Pois{\lambda_{ij}^{\mathsmaller{(t)}}} &\quad t &\in \{1,2\} .
\end{aligned}
\end{equation}
where $c_j$ is an additional (optional) parameter that effectively represents variance heterogeneity across columns of the observed matrix.



\subsection{Decomposition Strategies}

We present two factorization approaches for the non-centrality parameters of the DNCB distribution, $\lambda_1$ and $\lambda_2$.
These two approaches are guided by the Poisson factorization literature~\citep{gopalan15scalable}. The first,
\emph{matrix factorization}, is based on the CP decomposition for tensor factorizations, and the second,
\emph{matrix tri-factorization}, is based on the Tucker decomposition.


\paragraph{Matrix Factorization}
DNCB matrix factorization (DNCB-MF) assumes $\lambdaijr$ and $\lambdaijg$ are linear functions of low-rank latent factors,
\begin{equation}
\label{eq:dncbmf}
    \lambdaijr = \sum_{k=1}^{K} \theta_{ik}^{\mathsmaller{(1)}}\phi_{kj} \quad \text{and} \quad \lambdaijg = \sum_{k=1}^{K} \theta_{ik}^{\mathsmaller{(2)}}\phi_{kj}.
\end{equation}
In the context of DNA methylation data, the vector of latent factors $\boldsymbol{\phi}_{j} \coloneqq \langle \phi_{1j} \dots \phi_{Kj} \rangle$ describes how relevant gene $j$ is to each of the $K$ latent components. The vectors $\boldsymbol{\theta}_i^{(1)} \coloneqq \langle \theta_{i1}^{\mathsmaller{(1)}} \dots \theta_{iK}^{\mathsmaller{(1)}} \rangle, \boldsymbol{\theta}_i^{\mathsmaller{(2)}} \coloneqq \langle \theta_{i1}^{\mathsmaller{(2)}} \dots \theta_{iK}^{\mathsmaller{(2)}} \rangle$ represent how methylated or unmethylated, respectively, genes in latent component $k$ are in sample $i$. The ratio $\rho_{ik} \coloneqq \frac{\theta_{ik}^{\mathsmaller{(1)}}}{\theta_{ik}^{\mathsmaller{(1)}} + \theta_{ik}^{\mathsmaller{(2)}}} \in (0, 1)$ indicates that pathway $k$ is hypermethylated in sample $i$ when $\rho_{ik} \gg 0.5$ and hypomethylated in sample $i$ when $\rho_{ik} \ll 0.5$. The vector $\boldsymbol{\rho}_{i} \in (0,1)^K$ can also be interpreted as an embedding of sample $i$.

\paragraph{Matrix Tri-Factorization}
DNCB matrix tri-factorization (DNCB-TD) assumes $\lambdaijr$ and $\lambdaijg$ factorize into three factor matrices
\begin{equation}
\label{eq:dncbtd}
    \lambda_{ij}^{(1)} = \sum_{c=1}^{C}\thetaic\sum_{k=1}^{K} \pick{1}\phi_{kj} \quad \text{and} \quad \lambda_{ij}^{(2)} = \sum_{c=1}^{C}\thetaic\sum_{k=1}^{K} \pick{2}\phi_{kj}.
\end{equation}
The vector of sample factors $\boldsymbol{\theta}_{i} \coloneqq \langle \theta_{i1} \dots \theta_{iC}\rangle$ describes how much sample $i$ belongs to each of $C$ overlapping \emph{sample clusters}.  Similarly, the vector of gene factors $\boldsymbol{\phi}_{j} \coloneqq \langle \phi_{1j} \dots \phi_{Kj}\rangle$ describes how much gene $j$ belongs to each of $K$ overlapping \emph{pathways} or \emph{feature clusters}.  The vectors $\boldsymbol{\theta}_{i}$ and $\boldsymbol{\phi}_{j}$ can be viewed as embeddings of sample $i$ and gene $j$, respectively. The core matrices $\pick{1}, \pick{2}$ connect the latent factors $\boldsymbol{\theta}_{i}, \boldsymbol{\phi}_{j}$ by describing the extent to which samples in cluster $c$ have genes that are hypermethylated in pathway $k$.

DNCB-TD decouples the ranks of the factor matrices, allowing the number of latent factors in $\thetaic$ to differ from that of $\phi_{kj}$. 
This offers a more flexible representation that is useful in applications where $I, J$ are different orders of magnitude, as in methylation data.
Both DNCB-MF and DNCB-TD are forms of 2-mode Tucker decomposition \citep{hoff2005bilinear,nickel2012factorizing,li2009non,tucker64extension}. 
DNCB-MF is a special case of Tucker decomposition, where the cardinalities of the latent factors $\theta_{ik}^{\mathsmaller{(t)}}, \phi_{kj}$ are equal. 


\paragraph{Priors}



To fully specify our Bayesian hierarchical model, the latent factors are endowed with Gamma priors giving
\begin{equation}
\label{eq:dncbmf-prior}
\begin{aligned}
    \theta_{ik}^{\mathsmaller{(1)}}, \theta_{ik}^{\mathsmaller{(2)}} &\sim \Gam{\eta_1, \eta_2} \\
    \phi_{kj} &\sim \Gam{\nu_1, \nu_2} \\
\end{aligned}
\end{equation}
for DNCB-MF, and
\begin{equation}
\label{eq:dncbtd-prior}
\begin{aligned}
    \thetaic &\sim \Gam{\eta_1, \eta_2} \\
    \phi_{kj} &\sim \Gam{\nu_1, \nu_2} \\
    \pick{1}, \pick{2} &\sim \Gam{\zeta_1, \zeta_2}
\end{aligned}
\end{equation}
for DNCB-TD.

\paragraph{Generative Model}

The following algorithm can be used to draw samples from the generative DNCB-MF model:
\begin{enumerate}
    \setlength{\itemsep}{0em}
    \item Sample from $\phi_{kj} \overset{i.i.d.}{\sim} \Gam{\nu_1, \nu_2} \text{ for } k=1, \ldots,K, \text{ and } j=1, \ldots, J$.
    \item Sample from $\theta_{ik}^{\mathsmaller{(1)}}, \theta_{ik}^{\mathsmaller{(2)}} \sim \Gam{\eta_1, \eta_2} \text{ for } i=1, \ldots, I, \text{ and } k=1, \ldots,K$.
    \item Compute $\lambda_{ij}^{\mathsmaller{(t)}} = \sum_{k=1}^{K} \theta_{ik}^{\mathsmaller{(t)}}\phi_{kj} \text{ for } t=1,2, \ i=1, \ldots, I, \text{ and } j=1, \ldots,J.$
    \item Sample from $y_{ij}^{\mathsmaller{(t)}} \sim \Pois{\lambda_{ij}^{\mathsmaller{(t)}}} \text{ for } t=1,2, \ i=1, \ldots,I, \text{ and } j=1, \ldots, J$.
    \item Sample from $\gamma_{ij}^{\mathsmaller{(t)}} \Big| y_{ij}^{\mathsmaller{(t)}} \sim \Gam{\eps{t} + y_{ij}^{\mathsmaller{(t)}}, c_j} \text{ for } t=1,2, \ i=1, \ldots,I, \text{ and } j=1, \ldots, J$.
    \item Compute $\beta_{ij} \Big| \gammaijr, \gammaijg  = \frac{\gammaijr} {\gammaijr + \gammaijg} \text{ for } t=1,2, \ i=1, \ldots,I, \text{ and } j=1, \ldots, J$.
\end{enumerate}
The following algorithm can be used to draw samples from the generative DNCB-TD model:
\begin{enumerate}
    \setlength{\itemsep}{0em}
    \item Sample from $\pick{1}, \pick{2} \sim \Gam{\zeta_1, \zeta_2} \text{ for } c=1, \ldots,C, \text{ and } K=1, \ldots, K$.
    \item Sample from $\phi_{kj} \sim \Gam{\nu_1, \nu_2} \text{ for } k=1, \ldots,K, \text{ and } j=1, \ldots, J$.
    \item Sample from $\thetaic \sim \Gam{\eta_1, \eta_2} \text{ for } i=1, \ldots,I, \text{ and } c=1, \ldots, C$.
    \item Compute $\lambda_{ij}^{\mathsmaller{(t)}} = \sum_{c=1}^{C}\thetaic\sum_{k=1}^{K} \pick{t}\phi_{kj} \text{ for } t=1,2, \  i=1, \ldots,I, \text{ and } j=1, \ldots, J$.
    \item Sample $y_{ij}^{\mathsmaller{(t)}} \sim \Pois{\lambda_{ij}^{\mathsmaller{(t)}}}, \text{ for } t=1,2, \ i=1, \ldots,I, \text{ and } j=1, \ldots, J$.
    \item Sample $\gamma_{ij}^{\mathsmaller{(t)}} \Big| y_{ij}^{\mathsmaller{(t)}} \sim \Gam{\eps{t} + y_{ij}^{\mathsmaller{(t)}}, c_j}, \text{ for } t=1,2, \  i=1, \ldots,I, \text{ and } j=1, \ldots, J$.
    \item Compute $\beta_{ij} \Big| \gammaijr, \gammaijg  = \frac{\gammaijr} {\gammaijr + \gammaijg} \text{ for } i=1, \ldots,I, \text{ and } j=1, \ldots, J$.
\end{enumerate}

\section{MCMC Inference}
\label{sec:inference}
\begin{figure}[h!]
     \centering
     \begin{subfigure}[b]{0.45\textwidth}
         \centering
         \includegraphics[width=\textwidth]{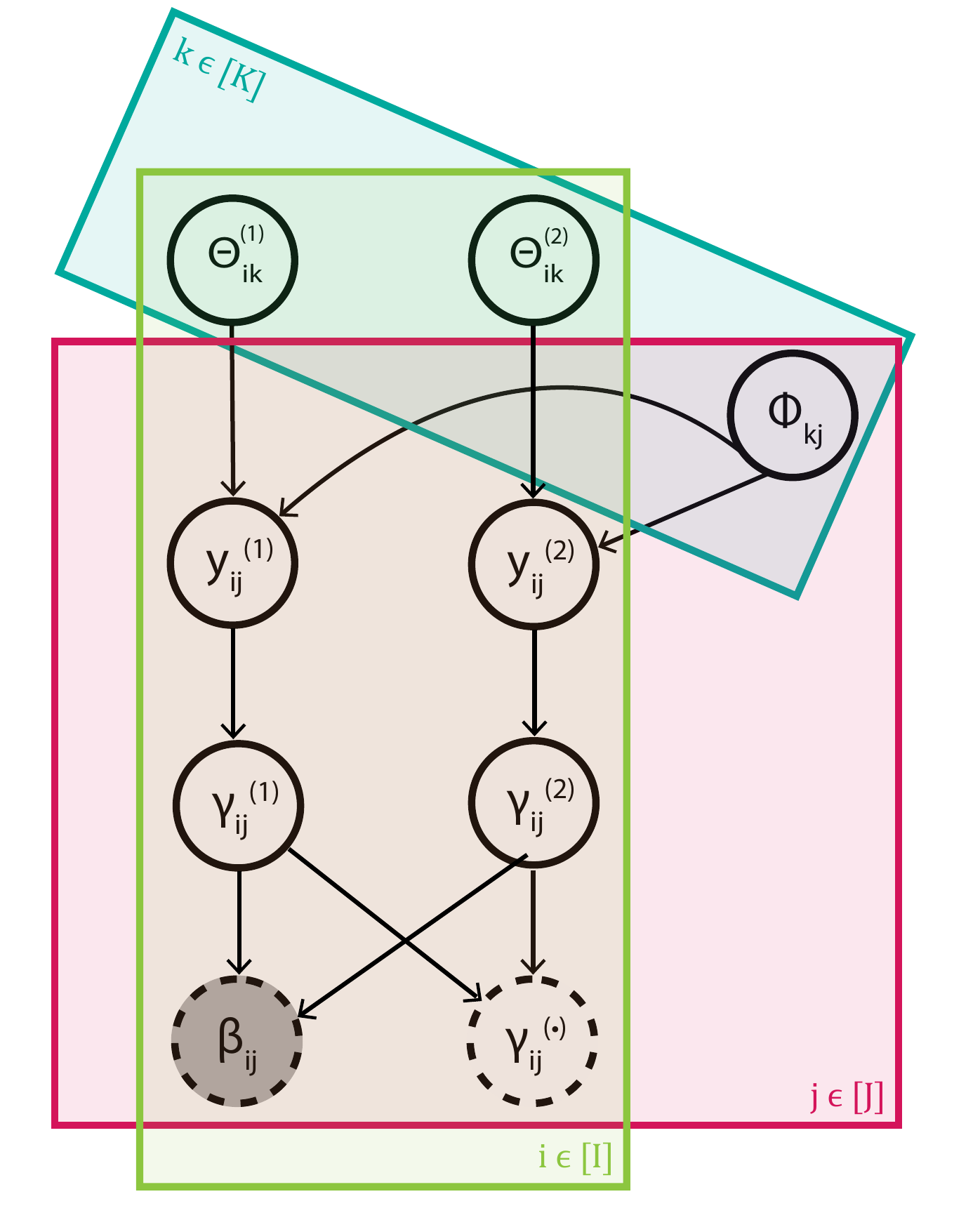}
         \caption{Graphical model representation of the fully augmented form of DNCB-MF given in \cref{eq:augmentedlikelihood,eq:dncbmf,eq:dncbmf-prior}.}
         \label{fig:dncbmf-aug}
     \end{subfigure}\hspace{1em}%
     \begin{subfigure}[b]{0.45\textwidth}
         \centering
         \includegraphics[width=\textwidth]{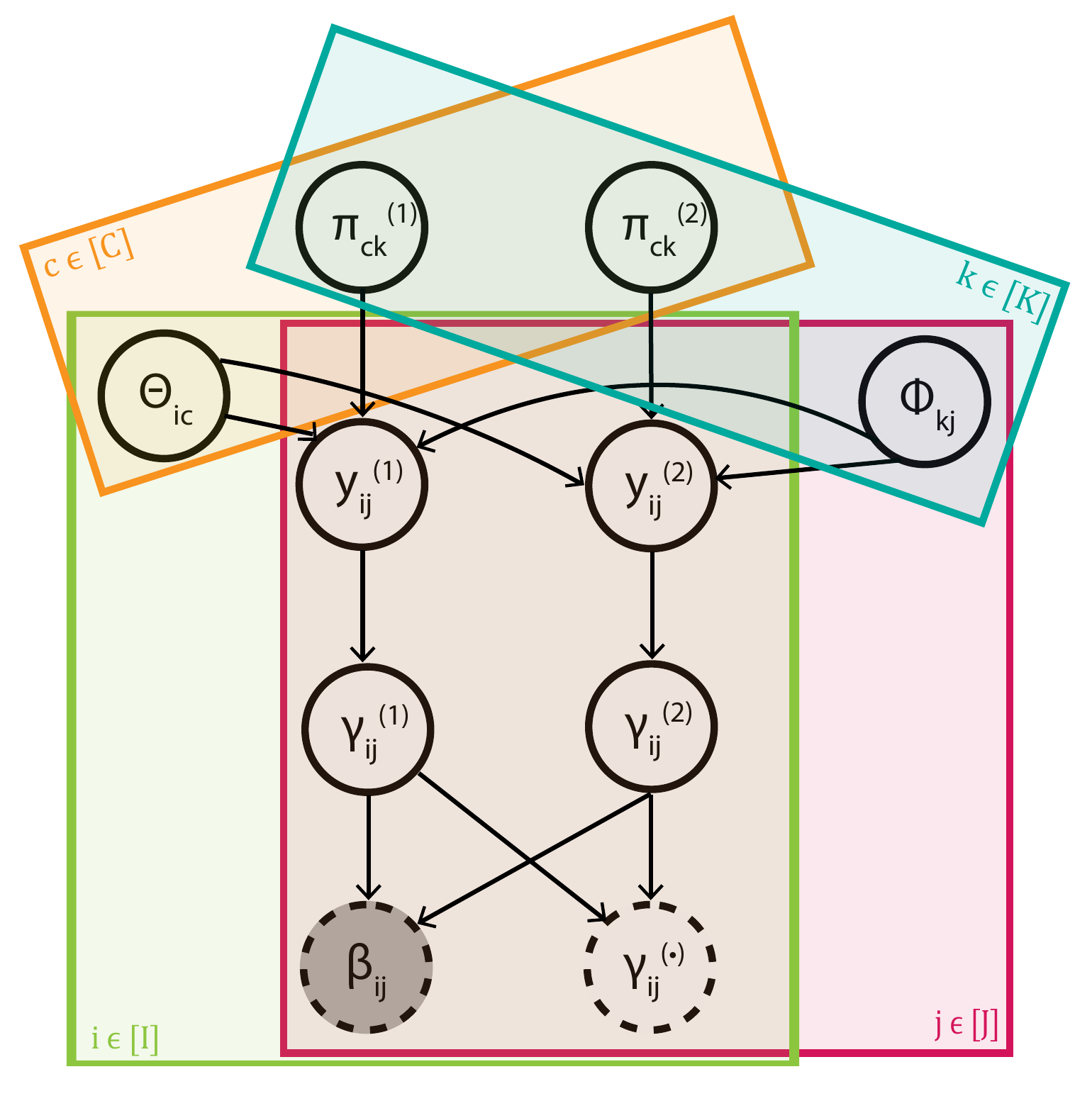}
         \caption{Graphical model representation of the fully augmented form of DNCB-TD given in \cref{eq:augmentedlikelihood,eq:dncbtd,eq:dncbtd-prior}.}
         \label{fig:dncbtd-aug}
     \end{subfigure}
    \caption{A graphical comparison of the DNCB-MF and DNCB-TD
          generative processes. The plate notation represents exchangeability across the specified indices. Shaded nodes are observed variables; unshaded nodes are latent variables. Solid edges denote random variables; dotted edges denote deterministic variables.}
        \label{fig:graphical-models}
\end{figure}

In this section, we provide the conditional posteriors for all the latent variables which collectively define a Gibbs sampler that is straightforward to implement and asymptotically guaranteed to generate samples from the posterior. Despite the lack of conjugacy to the beta likelihood (\ref{eq:betalikelihood}), we show that, under the augmented likelihood (\ref{eq:augmentedlikelihood}), the conditional posteriors for the counts $\yijm$ and $\yiju$ are available in closed form via the Bessel distribution \citep{yuan2000bessel}.
Graphical model representations of the augmented DNCB-MF and DNCB-TD decompositions are shown in \Cref{fig:graphical-models}.

\subsection{Sampling $\gamma_{ij}^{(1)}$ and $\gamma_{ij}^{(2)}$} 
The latent variables, $\gammaijr$ and  $\gammaijg$,  may be conditioned on and treated as data, if available.  
However, when only $\beta_{ij}$ is observed, we can sample the latent variables. 
We do so by appealing to the property given in Definition~\ref{df:prop-sum}.


\begin{definition}[Proportion-sum independence of gammas \citep{lukacs1955characterization}] 
\label{df:prop-sum}
If $\gamma_1 \sim \Gam{\alpha_1, c}$ and $\gamma_2\sim \Gam{\alpha_2, c}$ are independent gamma random variables with shared rate parameter $c$, then their sum $\gamma_{\cdot} \coloneqq \gamma_1 + \gamma_2$ and proportion $\beta \coloneqq \gamma_1 / \gamma_{\cdot} $ are \emph{independent} random variables:
$$\gamma_{\cdot}  \sim \Gam{\alpha_1 + \alpha_2, c}, \hspace{3em} \beta \sim \text{Beta}(\alpha_1, \alpha_2)$$
This holds if, and only if, $\gamma_1$ and $\gamma_2$ are gamma distributed.
\end{definition}

Since $\beta_{ij}$ is already defined as a proportion of the latent variables, we can redefine the latent variables as functions of their sum $\gamma^{(\cdot)}_{ij} \coloneqq \gammaijr + \gammaijg$,
\begin{equation}
\label{eq:redefine}
\gammaijr \coloneqq \beta_{ij}\gamma^{(\cdot)}_{ij},\hspace{4em} \gammaijg \coloneqq (1-\beta_{ij})\gamma^{(\cdot)}_{ij}.
\end{equation}
This implies that $\gamma^{(\cdot)}_{ij}$ is a gamma random variable,
\begin{equation}
\label{eq:magic}
\gamma^{(\cdot)}_{ij} \sim \Gam {\eps{1} + \eps{2} + y_{ij}^\mathsmaller{{(1)}} + y_{ij}^\mathsmaller{{(2)}}, \, c_j},
\end{equation}
which, by Definition~\ref{df:prop-sum}, is \emph{independent of the data} $\beta_{ij}$.  We can therefore sample it from the prior, as in \Cref{eq:magic} and then set $\gammaijr$ and $\gammaijg$ to the values given in \Cref{eq:redefine}.

\subsection{Sampling $\yijm, \yiju$}

As shown by \citet{yuan2000bessel}, if $\gamma \sim \textrm{Gam}(b+ y, c)$ is a gamma random variable with rate $c$ and shape $b + y$, and $y \sim \text{Pois}(\zeta)$ is a Poisson random variable with rate $\zeta$, then the conditional posterior over $y$ follows the Bessel distribution, so
$$P(y|\gamma, c, b, \zeta) = \text{Bes}(y; b{-}1,\; 2\sqrt{c\gamma \zeta}).$$ 
We can therefore sample $\yijm$ from its conditional posterior,
\begin{equation}
\label{eq:bessel}
\compcond{\yijmm} \sim \textrm{Bes}\left(\eps{1} \tminus1, \, 2\sqrt{c_j \gammaijr \lambdaijr}\right).
\end{equation}
The equation for $\yiju$ is the same but with $\eps{2}$ and $\lambdaijg$, rather than $\eps{1}$ and $\lambdaijr$.

\paragraph{The Bessel distribution.} Since the Bessel distribution is so uncommon, we take some space here to define some of its key properties. A Bessel random variable $y \sim \text{Bes}(v, a)$ is discrete $y \in \{0,1,2,\dots\}$.  The distribution has two parameters $v > -1$  and $a > 0$ and its unnormalized probability mass function is~\looseness=-1
\begin{equation}
\text{Bes}(y; v, a) \propto \frac{1}{y!\Gamma(y+v)}\Big(\frac{a}{2}\Big)^{2y+v}.
\end{equation}
The distribution is named for its normalizing constant
\begin{equation}
I(v,a) = \sum_{n=0}^\infty \frac{1}{n!\Gamma(n+v)}\Big(\frac{a}{2}\Big)^{2n+v}
\end{equation}
which is the modified Bessel function of the first kind.
A Bessel random variable's mean and variance are
\begin{align}
\mathbbm{E}[Y|v,a] &= \mu = \frac{a R(v, a)}{2} \\
\textrm{Var}(Y|v, a) &= \mu\big(1 + \mu (R(v+1, a) - R(v, a))),
\end{align}
where $R(v, a)\coloneqq \frac{I(v+1, a)}{I(v, a)}$ is called the Bessel quotient. Since the Bessel quotient is strictly decreasing in $v$ \citep{devroye2002simulating}, the quantity $R(v+1, a) - R(v, a)$ in the expression for the variance is always negative.  As a result, the Bessel distribution is underdispersed since its variance divided by its mean is upper-bounded by 1:~\looseness=-1
\begin{equation}
\frac{\textrm{Var}(Y|v, a)}{\mu} = 1 + \mu \big(R(v+1, a) - R(v, a)\big) \,\,\leq 1
\end{equation}

There does not currently exist any open-source implementations of Bessel sampling.  One contribution of this paper is an open-source Cython library that provides fast Bessel sampling methods.  Our library implements the four  rejection samplers of \citet{devroye2002simulating}. Some of these methods rely on direct computation of the Bessel function, which is available in the GNU Standard Library.  Others avoid calling the Bessel function---which can be numerically unstable---by relying entirely on computing the Bessel quotient, which can be computed without any special functions using the dynamic program of \citet{amos1974computation}.  We also implement the Gaussian approximation method of \citet{yuan2000bessel} and the table sampling method based on direct evaluation of the PMF, used by \citet{zhou2015gamma}.~\looseness=-1

\subsection{Sampling $y^{(\textrm{1})}_{ijck}, y^{(\textrm{2})}_{ijck}$} Due to the additive property of Poisson random variables \citep{kingman72poisson}, the count $y_{ij}^{(t)}$ can be defined as the sum of subcounts, $y_{ij}^{(t)} {\coloneqq} \sum_{c=1}^C\sum_{k=1}^K y^{(t)}_{ijck}$, each of which are Poisson:
\begin{equation}
 y^{(\textrm{t})}_{ijck} \stackrel{\text{ind.}}{\sim}\text{Pois}( \theta_{ic}\pick{t} \phi_{kj}) \;\;\;\forall \; c, k \textrm{ and } t \in \{1, 2\}
\end{equation}
These latent subcounts are required as sufficient statistics to compute the conditional posteriors over $\boldsymbol{\theta}_i, \boldsymbol{\phi}_j,\pick{t}, \text{ and }\zeta$.  Conditioned on their sum, the posterior over $\{ \yicjkmm \}_{c,k}$ is multinomial \citep{steel1953relation},
\begin{equation}
\label{eq:multi}
\compcond{\big\{\yicjkmm \big\}_{c,k}} \sim \textrm{Mult}\Big(\yijmm, \Big\{\frac{\theta_{ic}\pick{1}\phi_{kj}}{\rho_{ij}}\Big\}_{c,k}\Big).
\end{equation}
As described by \citet{schein2016bayesian}, the multinomial probabilities can be computed efficiently by exploiting the compositional structure of the Tucker product, $\rho_{ij}=\boldsymbol{\theta}_i\Pi \boldsymbol{\phi}_j^T$.

\subsection{Sampling $\boldsymbol{\theta}_i, \boldsymbol{\phi}_j,\pi_{ck}^{(t)}$}

By gamma-Poisson conjugacy, we can sample

\begin{equation}
    \begin{aligned}
    \compcond{\theta_{ik}^{\mathsmaller{(t)}}} &\sim \Gam{\eta_1 + \sum_{j=1}^J y_{ijk}^{\mathsmaller{(t)}}, \eta_2 + \sum_{j=1}^J \phi_{kj}}, \textrm{ and } \\
    \compcond{\phi_{kj}} &\sim \Gam{\nu_1 + \sum_{i=1}^{I}\sum_{t=1}^{2}y_{ijk}^{\mathsmaller{(t)}}, \nu_2 + \sum_{i=1}^{I}\sum_{t=1}^{2}\theta_{ik}^{\mathsmaller{(t)}}}
    \end{aligned}
\end{equation}
for DNCB-MF and
\begin{equation}
    \begin{aligned}
    \compcond{\thetaic} &\sim \Gam{\eta_1 + \sum_{j=1}^{J}\sum_{k=1}^{K}\sum_{t=1}^{2}y_{icjk}^{\mathsmaller{(t)}}, \eta_2 + \sum_{j=1}^{J}\sum_{k=1}^{K}\sum_{t=1}^{2}\pick{t}\phi_{kj}}, \\
    \compcond{\phi_{kj}} &\sim \Gam{\nu_1 + \sum_{i=1}^{I}\sum_{c=1}^{C}\sum_{t=1}^{2}y_{icjk}^{\mathsmaller{(t)}}, \nu_2 + \sum_{i=1}^{I}\sum_{c=1}^{C}\sum_{t=1}^{2}\theta_{ic}\pick{t}}, \textrm{ and }  \\
    \compcond{\pick{t}} &\sim \Gam{\zeta_1 + \sum_{i=1}^I\sum_{j=1}^J y_{icjk}^{\mathsmaller{(t)}}, \zeta_2 + \sum_{i=1}^I \thetaic \sum_{j=1}^J \phi_{kj}}
    \end{aligned}
\end{equation}
for DNCB-TD.


\section{Predictability, Computability, and Stability}
\label{sec:pcs}
In many application areas, data analysis methods must be reliable, reproducible, and transparent.
The principles of predictability, computability, and stability (PCS) help to ensure that these goals are achieved~\citep{yu2020veridical}.
In this section, we apply the PCS framework to analyze the predictibility, computability, and stability of DNCB-MF and DNCB-TD.
To measure predictability, we imputed missing data values; to measure stability, we measure the co-clustering of samples and features while varying the hyper-parameters of the model; and to address computability, we report computational complexity. As DNCB-TD is designed to model matrix data where $I \ll J$, we focus on cases where $C < K$ in these tasks.
We are looking for methods that have good accuracy on held-out data, are stable to variations in model hyper-parameters, and are computable in a time scale that is appropriate for the problem, and we find that DNCB-TD achieves the veridical goals of the PCS framework.  

\paragraph{Microarray Methylation Data.} We compiled a dataset of 400 cancer samples from the Cancer Genome Atlas (TCGA)~\citep{tcga}.
The 400 samples consisted of four clusters of 100 samples from four different cancer types: breast, ovarian, colon and lung cancer. 
Of the 27,578 loci that are available, we considered only the 5,000 with the highest variance across as samples, as has been done previously~\citep{ma2014comparisons}.  
The resultant matrix of $\beta$ values was 400 $\times$ 5,000. We make the dataset we used available: \url{https://github.com/flahertylab/dncb-fac/tree/main/data/methylation}.

\paragraph{Bisulfite Methylation Data.} 
We downloaded the dataset studied by~\citet{sheffield2017dna}. 
This dataset consists of 156 Ewing sarcoma cancer samples and 32 healthy samples ($N \teq 188$), whose methylation was profiled using bisulfite sequencing
(bi-seq). 
Bi-seq data consists of binary ``reads'' of methylation at
many loci per gene. 
Following standard framework, we processed this data into ``beta values'' by first counting all the methylated-mapped reads $d_{ij}$ and non-methylated reads $u_{ij}$ for all loci within a given gene $j$ and then calculating $\beta_{ij} = \frac{s_0 + d_{ij}}{2s_0 + d_{ij} + u_{ij}}$ with the smoothing term set to $s_0 \teq 0.1$ \citep{betavalues}. 
As with the microarray data, we selected the 5,000 genes with the highest variance to obtain a $188 \times 5,000$ matrix.

\paragraph{Olivetti Faces Data.} We loaded the Olivetti faces dataset \citep{olivettifaces} from scikit-learn's dataset library. The dataset consists of ten distinct $64\times64$ images of each of 40 different subjects. The ten images of each subject vary in facial expression, angle, and lighting. Each image is in greyscale and quantized to floating point values on the interval $[0, 1]$. We subset to the first 20 subjects, selecting the three most similar photos of each subject measured by Euclidean distance, and vectorized each of the resulting $60$ images to create a $60 \times 4096$ matrix with entries between $0$ and $1$. We make this dataset available: \url{https://github.com/flahertylab/dncb-fac/tree/main/data/faces}.

\paragraph{BG-NMF Method.}
Our model is most closely related to beta-gamma
non-negative matrix factorization (BG-NMF), which was developed by \citet{ma2015variational} specifically for DNA methylation
datasets. BG-NMF is the first (and, to our knowledge, only) matrix factorization model to assume a beta likelihood. Specifically, it assumes that each element $\beta_{ij} \in (0,1)$ in a sample-by-gene matrix is drawn,
\begin{align}
    \beta_{ij} &\sim \textrm{Beta}\left(\alphaij{1},\alphaij{2}\right),
\end{align}
where the two shape parameters $\alphaij{1}$ and $\alphaij{2}$ are defined to be the same linear functions of low-rank latent factors as those given in \Cref{eq:dncbmf}. BG-NMF also places the same gamma priors over these factors as those given in \Cref{eq:dncbmf-prior}.

DNCB-MF and BG-NMF both factorize a sample-by-gene matrix into three non-negative latent factor
matrices; however, DNCB-MF factorizes the non-centrality parameters of the DNCB distribution, while BG-NMF factorizes the shape parameters of the beta distribution. Deriving an efficient and modular posterior inference algorithm for BG-NMF is hampered by the lack of a closed-form conjugate prior for the beta distribution. \citet{ma2015variational} propose a variational inference algorithm that maximizes nested lower bounds on the model evidence. Their derivation is sophisticated, but highly tailored to the specific structure of the model, which makes the model difficult to modify or extend. Moreover, the quality of this algorithm's approximation to the posterior distribution is not well understood. For biomedical settings, in which precise quantification of uncertainty is often necessary, the lack of an efficient MCMC algorithm therefore limits BG-NMF's applicability.

\subsection{Predictability}

\paragraph{Prior Predictive Check.}
The first condition of predictability in a Bayesian workflow is that the model be capable of producing data that reflects the statistics of the actual observed data \citep{cemgil09bayesian}. In order to test whether this criterion was met, we performed a prior predictive check. 
A set of simulated data was generated from the model prior to fitting and the distribution was compared to the distribution of the observed data. 

\begin{figure}
\centering
\includegraphics[width=\linewidth]{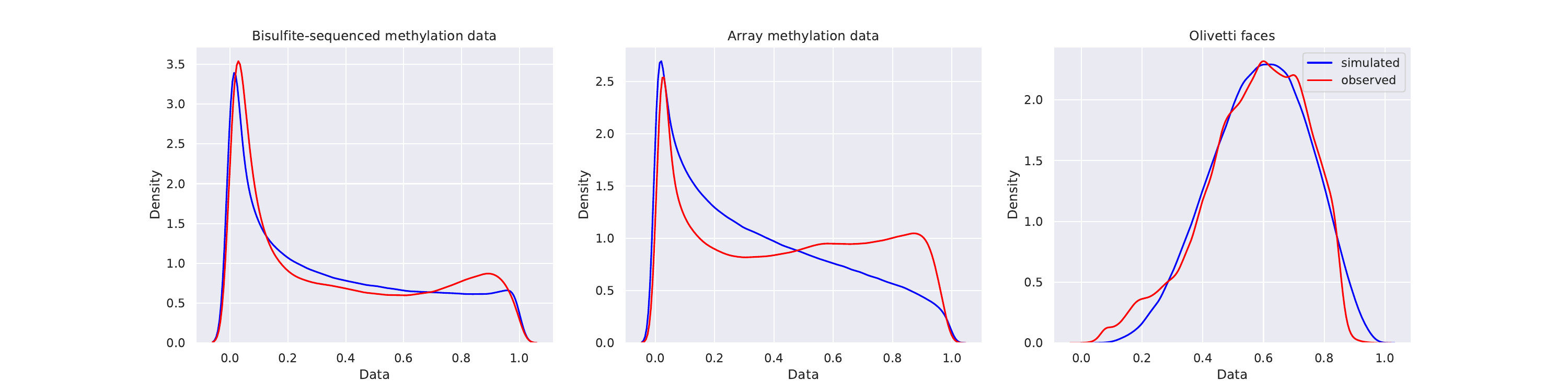}
\caption{Prior predictive checks for DNCB-TD on three datasets.}
\label{fig:predictability}
\end{figure}

Samples were drawn from the generative distribution for
DNCB-TD given in \Cref{eq:dncbtd-prior}, \Cref{eq:dncbtd}, and \Cref{eq:augmentedlikelihood}.
The similarity between the observed and simulated data was measured using mean squared error (MSE), $\frac{1}{N}\sum_{n=1}^N (\beta_{n}- \hat{\beta}_{n})^2$, for the $N$ total observed beta value entries $\beta_n$ and predictions $\hat{\beta}_n$. 
This process was repeated 1,000 times.
\Cref{fig:predictability,table:ppc-mse} contain results of the prior predictive analysis. The density plots indicate that 
DNCB-TD can produce data sets that are comparable to the methylation and Olivetti face datasets.

\begin{table}[h!]
\centering
\begin{tabular}{lc}\toprule
    \textbf{Dataset} & \textbf{MSE} \\\midrule
    Bisulfite sequenced methylation & $0.2000 \pm 0.0002 $ \\ 
    Array methylation & $0.1755 \pm 0.0001$ \\
    Olivetti faces & $0.0538 \pm 0.0001$ \\
    \bottomrule
\end{tabular}
\caption{Similarity between the observed and simulated data from DNCB-TD measured by mean squared error (MSE). MSE values are averaged over 1000 trials plus or minus one standard deviation.
}
\label{table:ppc-mse}
\end{table}

\paragraph{Heldout Prediction.} We randomly generated three masks that censored 10\% of the data matrix. 
Using the mask as input, we fit three models designed for bounded data (DNCB-TD, DNCB-MF, and BG-NMF) using three different random initializations and imputed the held-out values. 

To assess out-of-sample predictive performance, we used the pointwise predictive density (PPD)~\citep{gelman2014understanding}. For the three models, the PPD is given by
\begin{equation}
\label{eq:ppd}
    \textrm{PPD}_{\textrm{post}} = \prod_{i,j \in \mathcal{M}} \Big[\tfrac{1}{S} \sum_{s=1}^S P\big(\beta_{ij} \,|\, \Theta^{(1)}_s, \Theta^{(2)}_s, \Phi_s\big)\Big],
\end{equation}
where $\Theta^{(1)}_s, \Theta^{(2)}_s, \Phi_s$ are samples from the posterior distribution, either saved during MCMC for DNCB-MF and DNCB-TD or drawn from the fitted variational distribution for BG-NMF. For both models, we used $S\teq 100$. The predictive density $P(\cdot|\cdot)$ is the
beta distribution for BG-NMF and the DNCB distribution for DNCB-MF and DNCB-TD. 

\begin{figure}
\centering
\includegraphics[width=.8\linewidth]{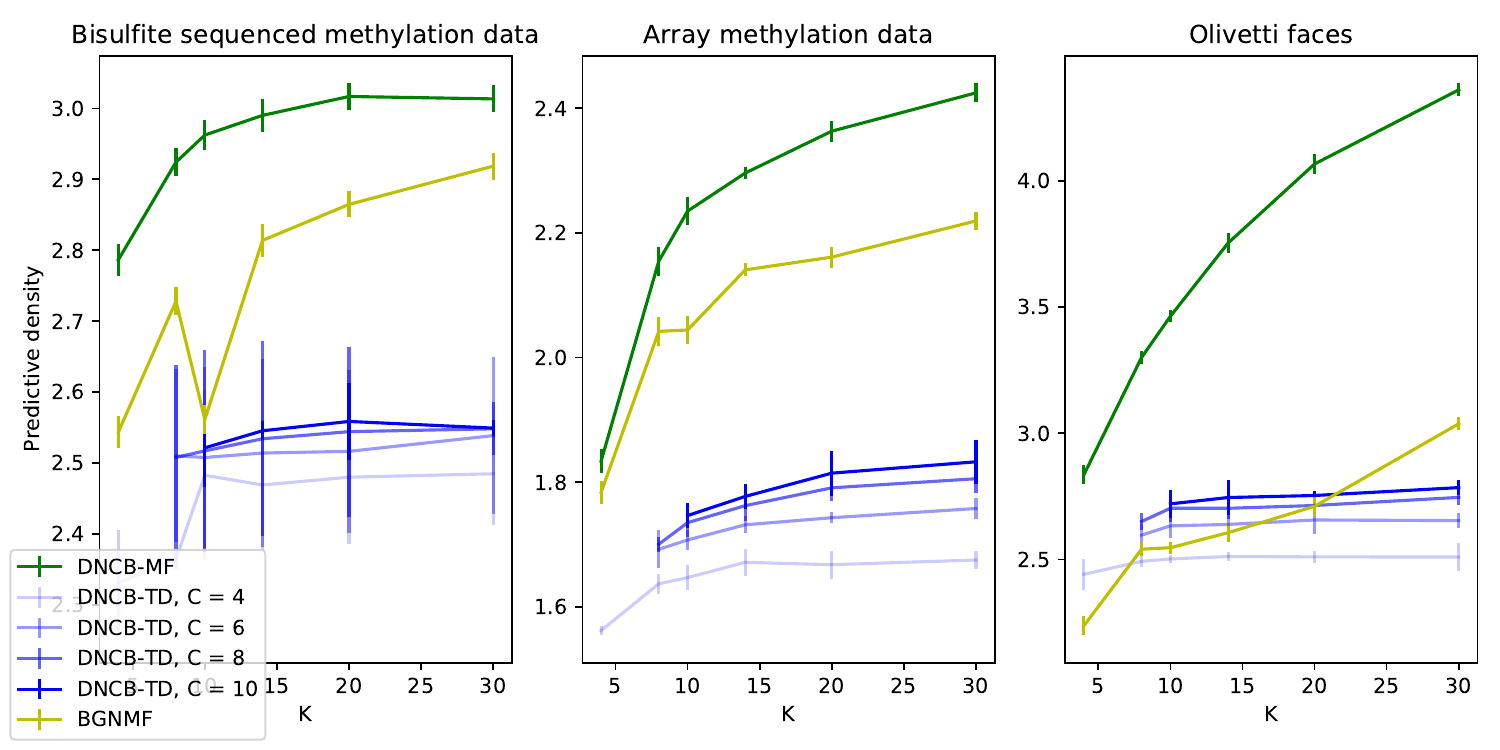}
\caption{Heldout prediction results on three datasets; higher is better. Random test-train splits were generated by creating three binary masks, each holding out a random 10\% of the data. Using three random initializations for each model, we fit the models on the training data and imputed the held-out values, varying K across 6 values. We plot the rescaled pointwise predictive density (PPD) obtained by each model; the error bars denote 95\% confidence intervals. All three models perform comparably well on held-out prediction.}
\label{fig:heldout}
\end{figure}

Figure~\ref{fig:heldout} shows $\textrm{PPD}^{\frac{1}{|\mathcal{M}|}}$, where $|\mathcal{M}|$ is the number of held-out values, varying the factor matrix cardinality, $K$. 
The performance metric is equivalent to the geometric mean of the predictive densities across the held-out values and is therefore comparable across all experiments.
All three models perform comparably well on held-out prediction. 

While imputation tasks may be useful for revealing the ways in which different models' inductive biases differ, they are not the inferential focus of tasks in the problem domain of DNA methylation.  The real motivating task for statistical models of DNA methylation is unsupervised discovery of cancer subtypes and the pathways that are relevant to those subtypes \citep{laird2010principles}.
Towards this end, it is critical that the inferred subtypes and pathways are stable for small variations of model parameters.

\subsection{Stability}
Stability assesses how experimental outcomes are affected by human judgment calls in the modeling process. A stable model's results should not significantly change with reasonable perturbations to the data or model architecture~\citep{yu2020veridical}. A typical human judgment call in matrix factorization is the cardinality of the factor matrices. 
For DNCB-TD, cardinality is controlled by two parameters $C, K$, that respectively define the cardinality of the $\Theta_{IC}$ and $\Phi_{KJ}$ factor matrices; for DNCB-MF and BG-NMF, cardinality is controlled by $K$. 
It is important in cancer subtype detection that cluster and pathway assignments are stable to reasonable changes in cardinality. 

\begin{figure}[ht] 
\centering
\includegraphics[width=0.8\linewidth]{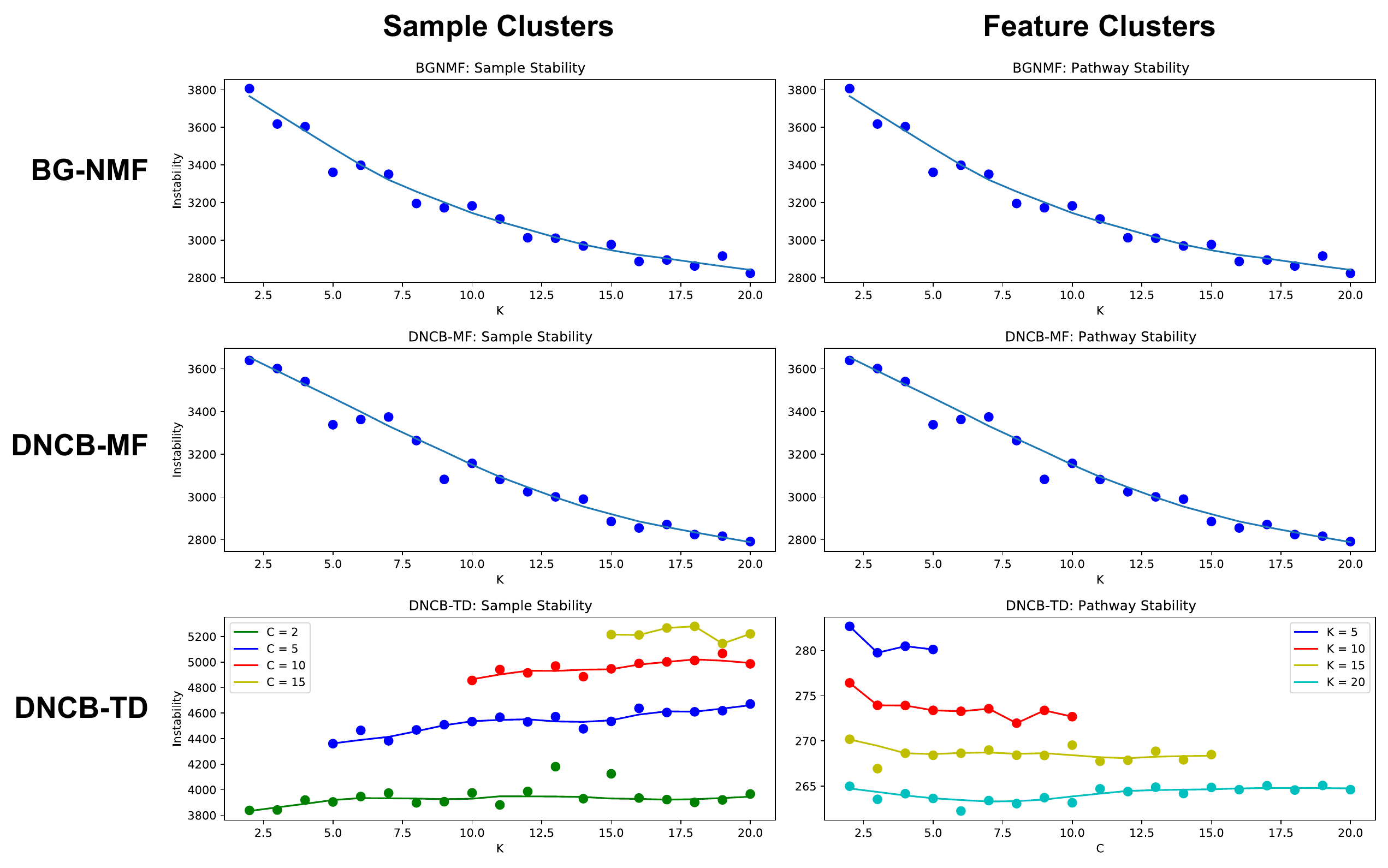}
\caption{Stability results for BG-NMF, DNCB-MF, and DNCB-TD on bisulfite sequencing methylation data. DNCB-TD is the only model for which the stability of both cluster and pathway assignments remains relatively constant as factor matrix cardinality increases.}
\label{fig:stability}
\end{figure}

For each combination of $C, K \in \{2, \ldots, 50\}$, DNCB-TD was fit on bisulfite-sequencing methylation data consisting of 156 Ewing sarcoma samples and 32 healthy samples. 
To measure the stability of cluster and pathway assignments to changes in factor matrix cardinality, we constructed sample and feature co-occurrence matrices for each combination of $C$ and $K$, and computed the KL divergence between the true and model-induced co-occurrences while varying one of $C, K$. 
We followed the same process for BG-NMF and DNCB-MF, varying the cardinality parameter $K$ across $\{2, \ldots, 50\}$. 
Figure~\ref{fig:stability} shows that DNCB-TD is the only model for which both cluster and pathway stability remains relatively constant across a range of $C$ and $K$.

\subsection{Computability}

The complexity of the algorithm is dominated by the Bessel and multinomial steps of equations~\ref{eq:bessel} and \ref{eq:multi}. The Bessel step scales linearly with the size of the data matrix $\mathcal{O}(2IJ)$.  The complexity of the multinomial step is $\mathcal{O}(CKN_{>0})$ where $N_{>0} \leq 2IJ$ is the number of non-zero $\yijmm, \yijuu$ counts.  For very sparse counts, this step may be faster than the Bessel step. In the worst case, the multinomial step is $\mathcal{O}(2IJCK)$.  
 How closely the model can fit the data is a direct function of the magnitude of the counts, because the concentration of the likelihood $\text{Beta}(\beta_{ij}; \; \eps{1} + \yijmm, \eps{2} + \yijuu)$ is proportional to $\yijmm + \yijuu$.
The likelihood is maximized by taking the counts to infinity while keeping their proportion fixed to $\beta_{ij}$. The more sparse the counts, the more efficient the process. The sparsity of the counts is affected by the magnitude of the Gamma priors, wherein very small values of the shape and rate parameters induce smaller counts. 
In practice, we initialize the Gibbs sampler by running several iterations of the BG-NMF method, which allows the model to rapidly acquire a coarse-grained representation of the data.

\section{Real Data Analysis}

In this section, we explore how the latent structures in the DNCB-TD model reveal methylation profiles associated with cancer subtypes. We also explore DNCB-TD inferences for the Olivetti faces image data set. The distribution of observed values within the bounded interval for methylation and image data is very different, yet the flexibility of the DNCB-TD model enables it to capture salient features.

\subsection{DNA Methylation Data}
 
We fit DNCB-TD to the Microarray Methylation Data described previously with $C=4$ and $K=6$.
We ran 2,000 iterations of MCMC inference.  
Due to label switching, one cannot average over samples, so the inferred latent variables values are from the last sample. 

\begin{figure}[h!]
    \centering
    \includegraphics[width=\textwidth]{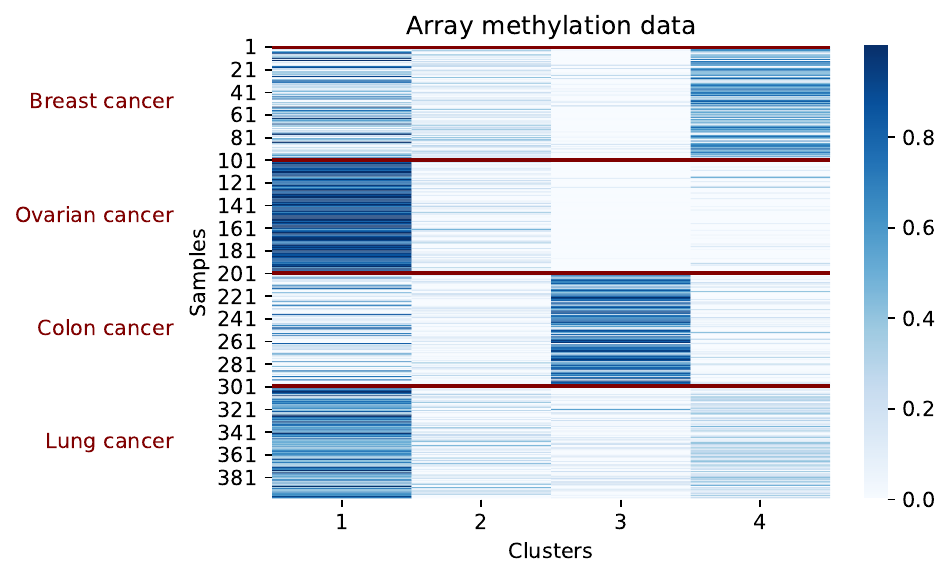}
    \caption{The inferred matrix $\hat{\Theta}$ for methylation array data. Rows are samples and columns are clusters---each entry represents the strength of association between sample $i$ and cluster $c$.}
    \label{fig:array-theta}
\end{figure}

DNCB-TD infers $C$ overlapping \emph{clusters} of samples.  
The extent to which sample $i$ is well-described by each of the $C$ clusters is given by the vector $\boldsymbol{\theta}_{i}$. 
Figure \ref{fig:array-theta} shows that DNCB-TD concentrates breast, ovarian, and colon samples in distinct clusters, indicating that the model infers structure that differentiates the cancer types. 

\begin{figure}[h!]
    \centering
    \includegraphics[width=.6\textwidth]{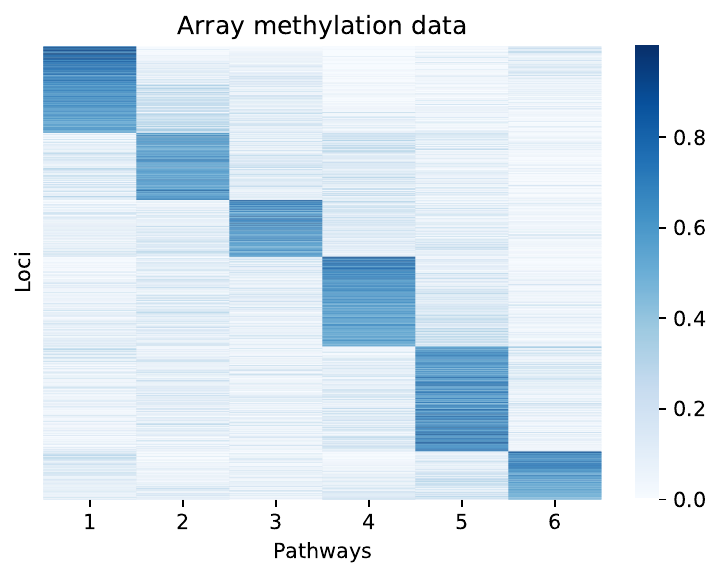}
    \caption{The inferred matrix $\hat{\Phi}$ for methylation array data. Rows are genes and columns are pathways---each entry represents the strength of association between pathway $k$ and gene $j$. Row indices are sorted according to pathway association.}
    \label{fig:array-phi}
\end{figure}

Similarly, our model infers $K$ overlapping \emph{pathways} of genes, and the extent to which gene $j$ is active in each of the $K$ pathways is given by the vector $\boldsymbol{\phi}_j$ (\Cref{fig:array-phi}). 
The core matrix of cluster-pathway factors $\Pi^{\left( 1 \right)} $ can be interpreted as a map between the space of clusters and pathways describing the extent to which samples in cluster $c$ methylate genes in pathway $k$. 

\begin{figure}[h!]
    \centering
    \includegraphics[width=.6\textwidth]{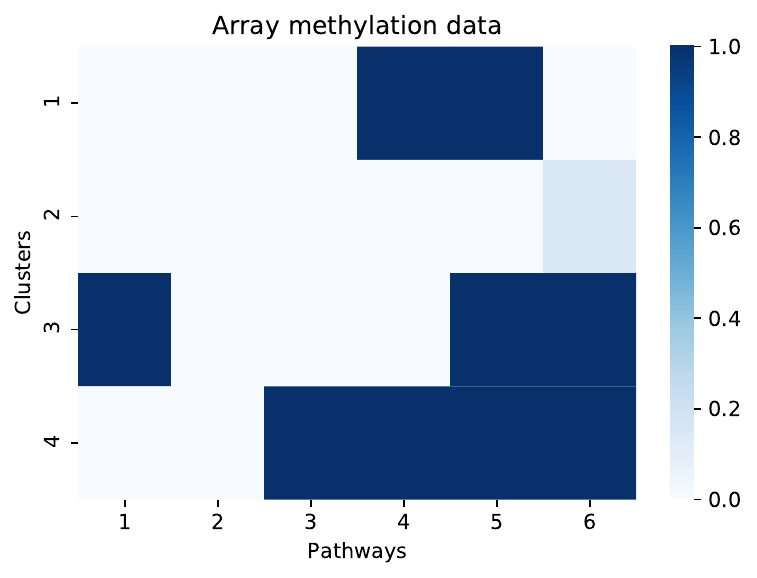}
    \caption{The inferred core matrix $\hat{\Pi}^{\left( 1 \right)}$ for methylation array data. Each entry represents the strength of association between samples in cluster $c$ and genes in pathway $k$.}
    \label{fig:array-pi}
\end{figure}

Figure \ref{fig:array-pi} illustrates a sparse, interpretable mapping of cancer types to genetic pathways. The learned pathways are consistent with cancer development mechanisms. 
Pathway 1 is strongly associated with sample cluster 3, which contains predominantly colon cancer samples; hypermethylation of genes KCNQ5 and ZNF625 in pathway 1 is a known biomarker of colorectal cancer \citep{kcnq5_2021,znf625_2015}.
Pathway 3 is associated with samples in sample cluster 4 which is composed primarily of breast cancer samples. A constituent of pathway 3 is the gene ASCL2 which has been shown to be associated with poor prognosis in breast cancer patients~\citep{ascl2}. 
Pathway 4 is associated with samples in sample clusters 1 and 4. Sample cluster 1 is composed of ovarian cancer samples among others. 
MUC13 is a component of pathway 4 and has been reported as a candidate biomarker for ovarian cancer detection~\citep{mucin13}.

A parallel analysis of the DNA methylation bisulfite sequencing data is in Appendix~\ref{app:bisulfite}.
The pathways/feature clusters show a clear block-diagonal structure similar to the array data.
Two sample clusters are characterized by absence of several pathways similar to Figure~\ref{fig:array-pi}.

\subsection{Olivetti Faces Data}
\begin{figure}[h!]
    \centering
    \includegraphics[width=.9\textwidth]{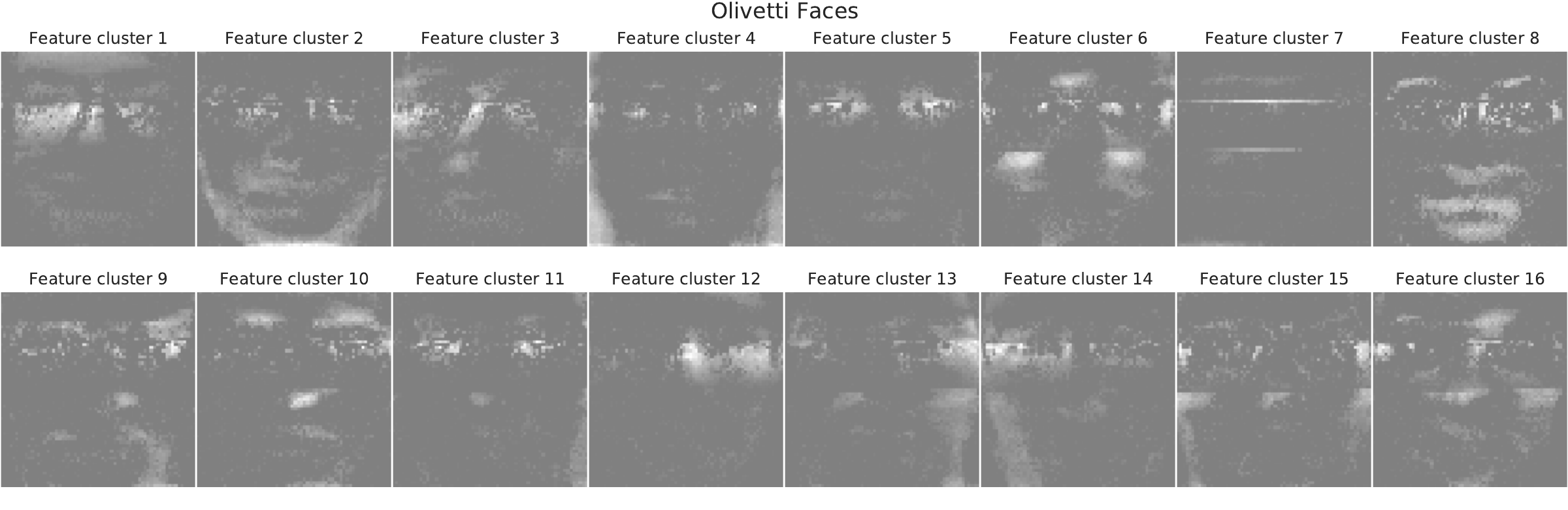}
    \caption{Inferred matrix $\hat{\Phi}$. Each feature cluster $\hat{\boldsymbol{\phi}}_k$ is converted to a $64 \times 64$ image.}
    \label{fig:olivetti-phi}
\end{figure}

We fit DNCB-TD to the Olivetti Faces Dataset with $C=20$ and $K=16$. We ran $5,000$ iterations of MCMC inference. The $K$ feature clusters are shown in Figure \ref{fig:olivetti-phi}. 
Clearly, the DNCB-TD representation is identifying salient groups of pixels that co-vary across samples. For example, feature cluster 8 seems to infer a correspondence between glasses and facial hair. These features are brighter pixel intensities indicating that the model is identifying a feature with a lack of glasses and facial hair. 

\begin{figure}[h!]
    \centering
    \includegraphics[width=\textwidth]{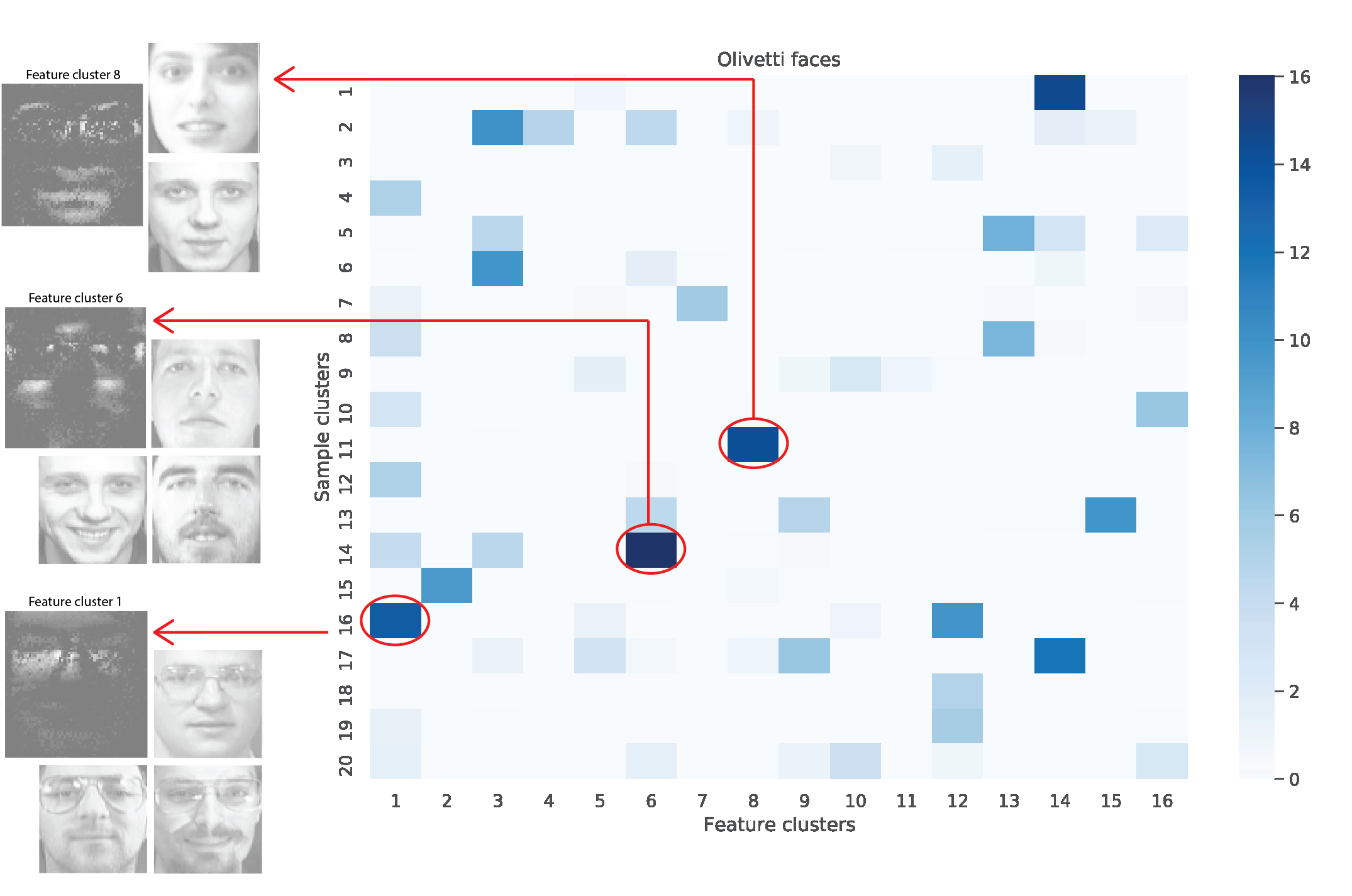}
    \caption{The inferred core matrix $\hat{\Pi}^{\left( 1 \right)}$ Olivetti faces data set. Each entry represents the strength of association between samples in cluster $c$ and features in pathway $k$.}
    \label{fig:olivetti-pi}
\end{figure}

Figure \ref{fig:olivetti-pi} shows that DNCB-TD concentrates individuals with similar features (eg. glasses) into distinct clusters that map interpretably to ``eigenfaces'' derived from the inferred matrix $\hat{\Phi}$.
Sample cluster 11 makes use of feature cluster 8. These samples and features correspond to individuals who have neither facial hair nor glasses.
Sample cluster 14 makes use of feature cluster 6. These samples and features correspond to individuals who have light pixels on the upper cheeks.
Sample cluster 16 makes use of feature cluster 1. These samples and features correspond to individuals who have glare on the glasses.

\begin{figure}[h!]
    \centering
    \includegraphics[width=\textwidth]{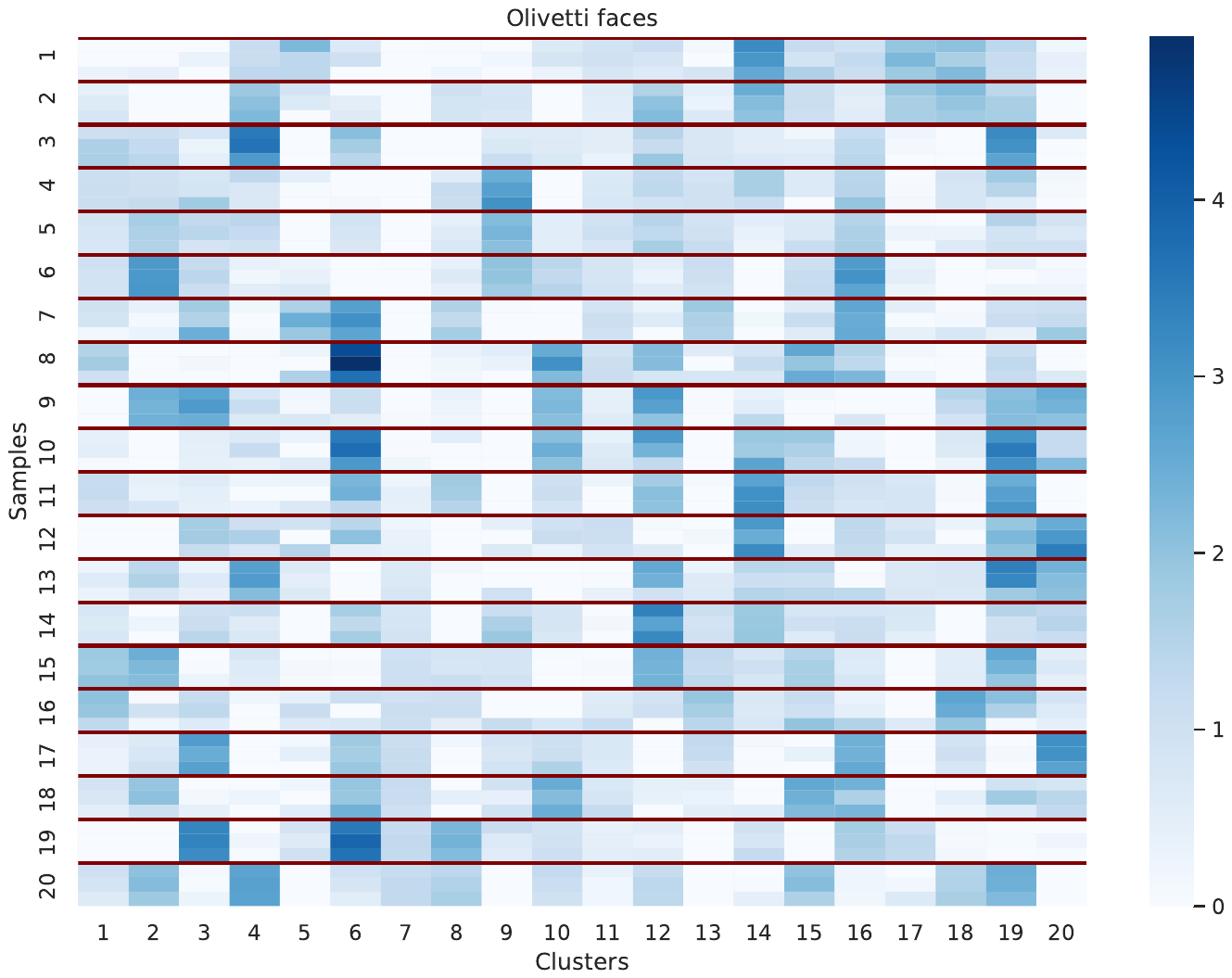}
    \caption{The inferred matrix $\hat{\Theta}$ for the Olivetti faces data set. Each entry represents the strength of association between sample $i$ and sample cluster $c$.}
    \label{fig:olivetti-theta}
\end{figure}

The assignment of samples to sample clusters in DNCB-TD is given by the inferred matrix $\hat{\Theta}$ shown in Figure~\ref{fig:olivetti-theta}. It is clear from the blocking structure that the DNCB-TD model is clustering samples from the same person into clusters.
Furthermore, samples from individuals that are similar fall into the same clusters.

\section{Conclusion}
Our goal in this work is to develop a predictive, stable, and computationally efficient method for matrix factorization of bounded support data.
One of the challenges of modeling general bounded support data is that there is a wide variety of empirical distributions of observed data.
By using the doubly non-central beta distribution as the likelihood distribution, we have achieved a level of flexibility that is required for general bounded support data. 

To model the observed data using latent factor representations, we proposed two methods: one based on the CP decomposition and one based on the Tucker decomposition.
The Tucker decomposition is more flexible than the CP decomposition because it allows the number of latent feature factors to be independently determined from the number of sample factors. 
CP decomposition and BG-NMF require these two dimensions to be the same.

The increased flexibility of the doubly non-central beta distribution and the Tucker decomposition come at an apparent cost - the distributions in the corresponding statistical model are not Bayesian conjugates and sampling may be computationally challenging. We show that by using the augment-and-marginalize trick, we are able to find closed form Gibbs sampling updates and mitigate the computational costs of the modeling choices.

We show that the model identifies informative latent factors in both DNA methylation data and image data.
In clinical and experimental applications, it is undesirable for a model to be sensitive to hyper-parameters and DNCB-TD is empirically stable to hyper-parameter changes. DNCB-TD has competitive held-out predictive performance as well, even though the primary objective of the model is to learn informative low-dimensional representations of the data.

Methylation patterns have the potential to improve our capability to use circulating cell-free DNA to monitor for cancer recurrence and localize metastatic events for further investigation.
The models proposed in this work could be used to identify informative biomarkers for those applications.
Indeed, matrix-valued data with bounded support are ubiquitous and there are many other applications of models for such data beyond clinical applications.


\acks{P.F. and A.N.A. acknowledge funding from NSF 1934846 and NIH 1R01GM135931-01. }


\newpage

\appendix
\section*{Appendix A.}

\subsection*{Real Data Analysis: Bisulfite sequencing methylation data}
\label{app:bisulfite}

We fit DNCB-TD with $C=3, K=8$ to a matrix of 188 bisulfite-sequenced samples, of which 156 were Ewing sarcoma cancer samples and 32 were healthy. We ran 2,000 iterations of MCMC inference. The resulting inferred matrices $\hat{\Theta}, \hat{\Phi}, \hat{\Pi}$ are visualized in Figures~\ref{fig:bisulfite-theta},~\ref{fig:bisulfite-phi},and ~\ref {fig:bisulfite-pi}. DNCB-TD is able to distinguish the Ewing sarcoma cell lines and tumors, and groups together the healthy and Ewing mesenchymal stem cells (MSCs).



\begin{figure}[h!]
    \centering
    \includegraphics[width=0.9\textwidth]{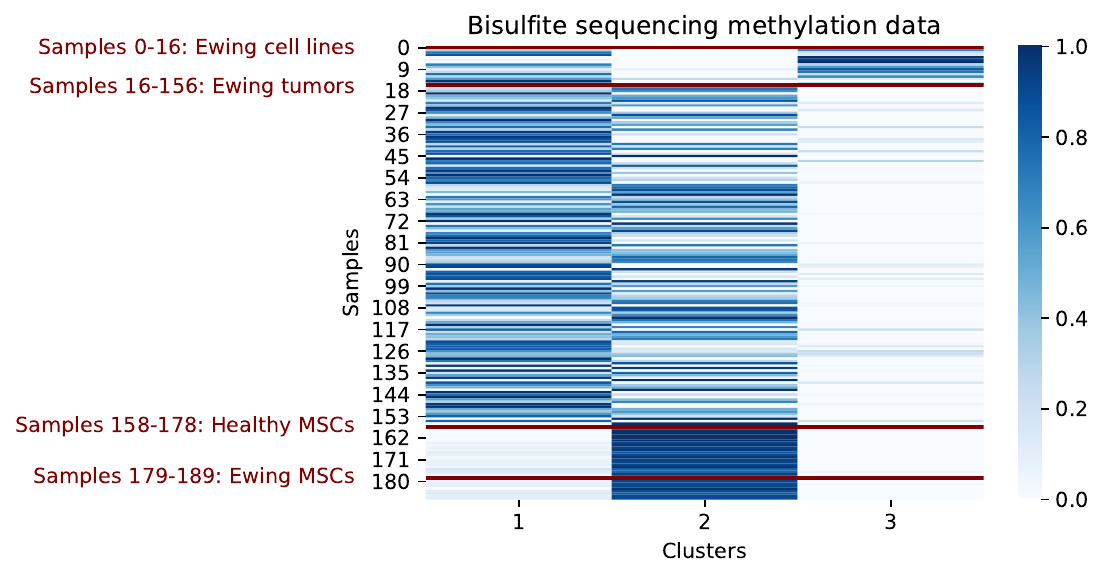}
    \caption{The inferred matrix $\hat{\Theta}$ for bisulfite methylation data. Rows are samples and columns are clusters---each entry represents the strength of association between sample $i$ and cluster $c$.}
    \label{fig:bisulfite-theta}
\end{figure}

\begin{figure}[htbp!]
    \centering
    \includegraphics[width=.9\textwidth]{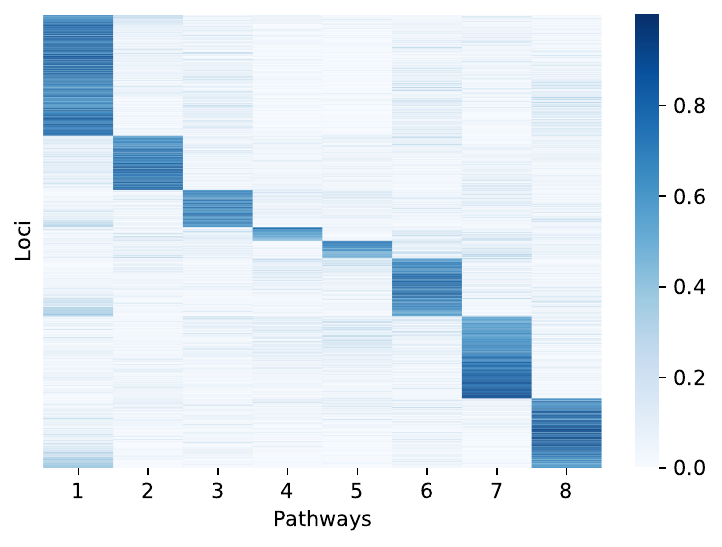}
    \caption{The inferred matrix $\hat{\Phi}$ for bisufite methylation data. Rows are genes and columns are pathways---each entry represents the strength of association between pathway $k$ and gene $j$. Row indices are sorted according to pathway association.}
    \label{fig:bisulfite-phi}
\end{figure}

\begin{figure}[h!]
    \centering
    \includegraphics[width=.9\textwidth]{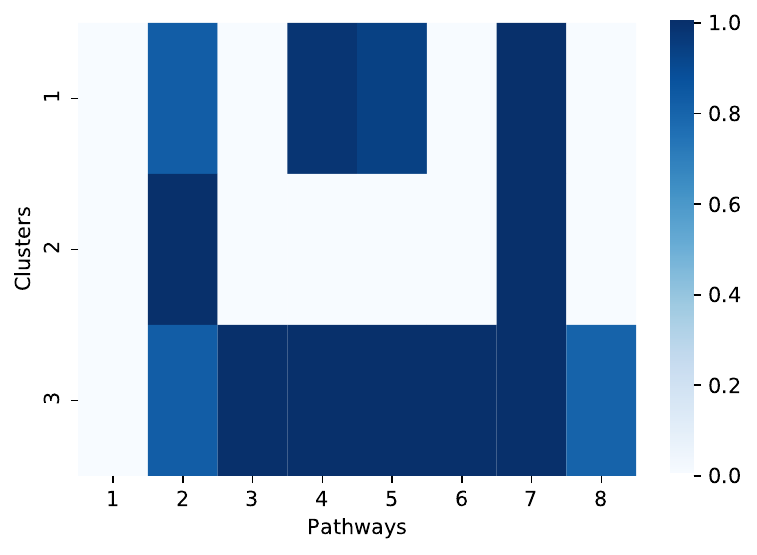}
    \caption{The inferred core matrix $\hat{\Pi}^{\left( 1 \right)}$ for bisulfite methylation data. Each entry represents the strength of association between samples in cluster $c$ and genes in pathway $k$.}
    \label{fig:bisulfite-pi}
\end{figure}

\clearpage

\vskip 0.2in
\bibliography{references}

\section{Proofs}
Conditioned on the local latent variables $\yijm,\yiju$, the complete likelihood is 
$$\beta_{ij}\sim \text{Beta}\left(b_0 + \yijm,\, b_0 + \yiju \right)$$.  The expectation of $\beta_{ij}$ under the complete likelihood is:
\begin{align}
\mathbb{E}\left[\beta_{ij}\,|\,b_0,\, \yijm, \yiju\right] &= \frac{b_0 + \yijm}{2b_0 + \yijm + \yiju}
\end{align}
From here on, we'll refer to this as $\mathbb{E}\left[\beta_{ij}\,|\,\yijm, \yiju\right]$, leaving implicit the conditioning on $b_0$.

We will refer to the expectation of $\beta_{ij}$ with the local latent variables integrated out simply as $\mathbb{E}[\beta_{ij}]$.  This leaves implicit the conditioning on $b_0, \zeta, \text{ and } \rho_{ij}$---i.e., $\mathbb{E}[\beta_{ij}] \equiv \mathbb{E}\left[\beta_{ij}\given b_0, \zeta, \rho_{ij}\right]$.  In general, we will leave implicit the conditioning on all model parameters other than $\yijm, \yiju$ (about which we will be explicit). By Law of Total Expectation, we have:
\begin{align}
\mathbb{E}\left[\beta_{ij}\right] &= \mathbb{E}_{\yijm,\yiju}\Big[\mathbb{E}\left[\beta_{ij}\,|\,\yijm, \yiju\right]\Big] \\
&= \mathbb{E}_{\yijm,\yiju} \left[\frac{b_0 +\yijm}{2b_0 + \yijm + \yiju}\right]
\intertext{We can then re-represent this in terms of $y_{ij}^{(\cdot)}\triangleq \yijm + \yiju$:}
&= \mathbb{E}_{\yijm, y^{(\cdot)}_{ij}} \left[\frac{b_0 +\yijm}{2b_0 + y^{(\cdot)}_{ij}}\right]
\intertext{Then again by Law of Total Expectation and linearity of expectation:}
&= \mathbb{E}_{y^{(\cdot)}_{ij}}\left[\mathbb{E}_{\yijm} \left[\frac{b_0 +\yijm}{2b_0 + y^{(\cdot)}_{ij}}\, |\, y^{(\cdot)}_{ij}\right]\right]\\
&= \mathbb{E}_{y^{(\cdot)}_{ij}}\left[\frac{b_0 + \mathbb{E}_{\yijm}\left[\yijm  | y^{(\cdot)}_{ij}\right]}{2b_0 + y^{(\cdot)}_{ij}}\right]
\intertext{Since $\yijm \sim \textrm{Pois}(\zeta\,\rho_{ij})$ we know that $\yijm|y_{ij}^{(\cdot)} \sim \textrm{Binom}(y_{ij}^{(\cdot)}, \rho_{ij})$.  Thus, the expectation is: $\mathbb{E}_{\yijm}\left[\yijm  | y^{(\cdot)}_{ij}\right] = \rho_{ij} \, y_{ij}^{(\cdot)}$, where recall that $\rho_{ij} = \boldsymbol{\theta}_i\,\Pi\,\boldsymbol{\phi}_j^T$ is computed from global variables only.  Plugging this in we get:}
&= \mathbb{E}_{y^{(\cdot)}_{ij}}\left[\frac{b_0 + \rho_{ij} \, y^{(\cdot)}_{ij}}{2b_0 + y^{(\cdot)}_{ij}}\right]
\end{align}

\begin{lemma} $\mathbb{E}\big[\beta_{ij}\big]$ has analytic closed form, equal to:\vspace{-0.5em}
$$\mathbb{E}\big[\beta_{ij}\big]= 0.5\,\textrm{M}(1,\, 2b_0\!+\!1,\, -\zeta) + \frac{\rho_{ij} \, \zeta}{2b_0\tplus1}\textrm{M}(1,2b_0\!+\!2,\tminus\zeta)$$
where $\textrm{M}(a,b,c)$ is Kummer's confluent hypergeometric function \citep{buchholz2013confluent}.
\end{lemma}

\textbf{Proof:} As established above, we have:
\begin{align}
\mathbb{E}[\beta_{ij}] &= \mathbb{E}_{y^{(\cdot)}_{ij}}\left[\frac{b_0 + \rho_{ij} \, y^{(\cdot)}_{ij}}{2b_0 + y^{(\cdot)}_{ij}}\right]
\intertext{Plugging in the Poisson probability mass function into the expectation, we get:}
&= \sum_{n=0}^\infty P\left(y^{(\cdot)}_{ij}\teq n\,|-\right)\left(\frac{b_0 + \rho_{ij} \, n}{2b_0 + n}\right) \\
&= \sum_{n=0}^\infty \textrm{Pois}\left(n;\,\zeta\right)\left(\frac{b_0 + \rho_{ij} \, n}{2b_0 + n}\right) \\
&= \sum_{n=0}^\infty \frac{\zeta^n \, e^{-\zeta}}{n!} \left(\frac{b_0 + \rho_{ij} \, n}{2b_0 + n}\right) \\
&= e^{-\zeta} \sum_{n=0}^\infty \frac{\zeta^n}{n!} \left(\frac{b_0 + \rho_{ij} \, n}{2b_0 + n}\right)
\end{align}
The infinite sum converges and has an analytic solution \citep{WolframAlpha}:
\begin{align}
\label{eq:wolfram}
&= e^{-\zeta} (-\zeta)^{-2b_0} \Bigg(b_0\,\Big(\Gamma(2b_0)-\Gamma(2b_0,\,-\gamma)\Big) - \rho_{ij}\, \Big(\Gamma(2b_0+1) -\Gamma(2b_0+1,-\gamma)\Big)\Bigg)
\end{align}
where $\Gamma(s, x)\teq\int_{x}^\infty \textrm{d}t \,t^{s\tminus1}e^{-t}$ is the upper incomplete gamma function. The upper incomplete gamma function is defined for negative values of its second argument; however almost all implementations suport only positive arguments.  As shown by \citet{gil2016computation}, the upper incomplete gamma function with a negative second argument can be defined as: 
\begin{align}
\Gamma(s, x) &= \Gamma(s)\Big(1 - x^s \, \gamma^*(s, x)\Big)
\intertext{where $\gamma^*(s, x) \teq \frac{x^{-s}}{\Gamma(s)} \gamma(s,x)$ is Tricomi's incomplete gamma function \citep{gautschi1998incomplete}, itself defined in terms of the lower incomplete gamma function $\gamma(s, x)=\int_0^x \text{d}t \, t^{s\tminus 1} e^{-t}$. The Tricomi's incomplete gamma function can then be defined as:}  
\gamma^*(s, x) &= \frac{e^{-x} \textrm{M}(1, s+1, x)}{\Gamma(s+1)}
\intertext{where $\textrm{M}(a,b,c)$ is Kummer's confluent hypergeometric function \citep{buchholz2013confluent}. We can therefore express the upper incomplete gamma function in terms of Kummer's confluent hypergeometric function, for which there are stable implementations that support negative arguments:}
\Gamma(s, x) &= \Gamma(s)\Big(1 - x^s \, \frac{e^{-x} \textrm{M}(1, s+1, x)}{\Gamma(s+1)}\Big)
\end{align}
Using this identity we will re-express equation~\ref{eq:wolfram} in terms of confluent hypergeometric functions. First, we rewrite the term $b_0\,\Big(\Gamma(2b_0)-\Gamma(2b_0,\,-\gamma)\Big)$:
\begin{align}
b_0\,\Big(\Gamma(2b_0)-\Gamma(2b_0,\,-\gamma)\Big) &= b_0\,\left(\Gamma(2b_0) -\Gamma(2b_0)\Big(1 - (-\zeta)^{2b_0} \, \frac{e^{\zeta} \textrm{M}(1, 2b_0+1, -\zeta)}{\Gamma(2b_0+1)}\Big)\right) \\ 
&= b_0\,\Gamma(2b_0)\,\left(1 - \Big(1 - (-\zeta)^{2b_0} \, \frac{e^{\zeta} \textrm{M}(1, 2b_0+1, -\zeta)}{\Gamma(2b_0+1)}\Big)\right) \\
&= b_0\,\Gamma(2b_0)\,\left((-\zeta)^{2b_0} \, \frac{e^{\zeta} \textrm{M}(1, 2b_0+1, -\zeta)}{\Gamma(2b_0+1)}\right) \\
&= \frac{b_0\,\Gamma(2b_0)}{\Gamma(2b_0+1)}\,(-\zeta)^{2b_0} \, e^{\zeta} \textrm{M}(1, 2b_0+1, -\zeta) \\
&= 0.5\,(-\zeta)^{2b_0} \, e^{\zeta} \textrm{M}(1, 2b_0+1, -\zeta)
\intertext{Next we rewrite the following term (using nearly identical steps):}
\rho_{ij}\, \Big(\Gamma(2b_0+1) -\Gamma(2b_0+1,-\gamma)\Big) &= \frac{\rho_{ij}\,\Gamma(2b_0+1)}{\Gamma(2b_0+2)}\,(-\zeta)^{2b_0+1} \, e^{\zeta} \textrm{M}(1, 2b_0+2, -\zeta) \\
&= \frac{\rho_{ij}}{2b_0+1}\,(-\zeta)^{2b_0+1} \, e^{\zeta} \textrm{M}(1, 2b_0+2, -\zeta)
\intertext{Plugging these two expressions back into equation~\ref{eq:wolfram} and then canceling some terms, we get:}
\label{eq:lemma1}
\mathbb{E}[\beta] = \mathbb{E}_{y^{(\cdot)}_{ij}}\left[\frac{b_0 + \rho_{ij} \, y^{(\cdot)}_{ij}}{2b_0 + y^{(\cdot)}_{ij}}\right] &= 0.5\,\textrm{M}(1, 2b_0+1, -\zeta) + \rho_{ij} \frac{\zeta}{2b_0+1}\,\textrm{M}(1, 2b_0+2, -\zeta)
\end{align}
which is in terms of two confluent hypergeometric functions that can be computed stably with a negative third argument in standard code libraries.\hfill $\blacksquare$

\begin{lemma}
For any $b_0 > 0$ and $\zeta > 0$, the following holds:\vspace{-0.5em}
$$\textrm{M}(1,\, 2b_0\!+\!1,\, -\zeta) \,+\, \frac{\zeta}{2b_0\tplus1}\textrm{M}(1,2b_0\!+\!2,\tminus\zeta) \,=\, 1.$$
Thus, $\mathbb{E}[\beta]$ is a linear function of $\rho_{ij}$ that can be written as:
$$\mathbb{E}[\beta] = 0.5\,q_{b,\zeta} - \rho_{ij}(1-q_{b,\zeta})$$
where we define $q_{b,\zeta} \triangleq \textrm{M}(1,\, 2b_0\!+\!1,\, -\zeta)$. 
\end{lemma}

\textbf{Proof:} The confluent hypergeometric function can be expressed as an infinite sum:
\begin{align}
\textrm{M}(a, b, c) = \sum_{n=0}^\infty \frac{a^{(n)}z^n}{b^{(n)}n!}
\end{align}
where $x^{(n)}$ denotes a rising factorial, equivalent to $x^{(n)} = \frac{\Gamma(x+n)}{\Gamma(x)}$.  When $a=1$ (as in our case), the rising factorial $a^{(n)}$ in the numerator equals $n!$ and cancels with the $n!$ term in the denominator.  If we also rewrite $b^{(n)}$ explicitly as a ratio of gamma functions, we get:
\begin{align}
\textrm{M}(1, b, c) = \sum_{n=0}^\infty \frac{z^n \Gamma(b)}{\Gamma(b+n)}
\end{align}
Proving the lemma reduces to showing that $\textrm{M}(1, b, c) - \frac{c}{b}\,\textrm{M}(1, b+1,c) = 1$, where $b=2b_0+1$ and $c=-\zeta$.  To show this, we represent the hypergeometric functions as infinite sums:
\begin{align}
\textrm{M}(1, b, c) - \frac{c}{b}\,\textrm{M}(1, b+1,c)
&= \sum_{n=0}^\infty \frac{c^n \Gamma(b)}{\Gamma(b+n)} - \frac{c}{b}\sum_{n=0}^\infty \frac{c^n \Gamma(b+1)}{\Gamma(b+1+n)} \\
&= \sum_{n=0}^\infty \frac{c^n \Gamma(b)}{\Gamma(b+n)} - \sum_{n=0}^\infty \frac{c^{n+1} \Gamma(b)}{\Gamma(b+1+n)} \\
&= \sum_{n=0}^\infty \frac{c^n \Gamma(b)}{\Gamma(b+n)} - \sum_{n=1}^\infty \frac{c^{n} \Gamma(b)}{\Gamma(b+n)} \\
&= 1 + \sum_{n=1}^\infty \frac{c^n \Gamma(b)}{\Gamma(b+n)} - \sum_{n=1}^\infty \frac{c^{n} \Gamma(b)}{\Gamma(b+n)} \\
& = 1
\end{align}
\hfill $\blacksquare$

\begin{theorem}
The model expectation of $\beta_{ij}$, with the local latent variables $\yijm, \yiju$ marginalized out, approaches $\rho_{ij}$ for small $b_0$ and large $\zeta$:
$\mathlarger{\lim_{b_0 \rightarrow 0}} \mathlarger{\lim_{\zeta \rightarrow \infty}}\mathbb{E}\big[\beta_{ij}\big] \rightarrow \rho_{ij}$.
\end{theorem}

\textbf{Proof:} With the result of Lemma 2, we can rewrite the expression for the expectation obtained in Lemma 1 simply as:
$$\mathbb{E}[\beta] = 0.5 \, q_{b,\zeta} + \rho_{ij} \,(1\tminus q_{b,\zeta})$$
where $q_{b,\zeta}=\textrm{M}(1, 2b_0+1, -\zeta)$.  From this it's easy to see that the expectation becomes $\rho_{ij}$ as $q_{b,\zeta}$ goes to 0: $\mathlarger{\lim}_{q_{b,\zeta} \rightarrow 0}\mathbb{E}[\beta] \rightarrow \rho_{ij}$.  Thus the proof reduces to showing:
$$\lim_{b_0\rightarrow 0} \lim_{\zeta \rightarrow \infty} \textrm{M}(1, 2b_0+1, -\zeta) \rightarrow 0$$
To show this, we first apply Kummer's transformation $\textrm{M}(a, b, c) = e^c \textrm{M}(b-a,b,-c)$:
\begin{align}
\lim_{b_0\rightarrow 0} \lim_{\zeta \rightarrow \infty}\textrm{M}(1, 2b_0+1, -\zeta) &= \lim_{b_0\rightarrow 0} \lim_{\zeta \rightarrow \infty}e^{-\zeta}\,\textrm{M}(2b_0, 2b_0+1, \zeta)
\intertext{We may then appeal to a limiting form of the hypergeometric function, $\textrm{M}(a, b, c)$ for when $c\rightarrow \infty$: $\textrm{M}(a, b, c) \sim e^c c^{a-b}/\Gamma(a)$.  Applying this transformation, we get:}
&= \lim_{b_0\rightarrow 0} \lim_{\zeta \rightarrow \infty} e^{-\zeta}\,e^{\zeta} \frac{\zeta^{2b_0-(2b_0+1)}}{\Gamma(2b_0)} \\
&= \lim_{b_0\rightarrow 0} \lim_{\zeta \rightarrow \infty}\, \zeta^{-1}\frac{1}{\, \Gamma(2b_0)} = 0
\end{align}
\hfill $\blacksquare$

\section{DNA Methylation Sequencing Data Analysis}
\label{sec:methyl-seq-analysis}

\begin{figure}[h!]
\caption{Prior Predictive Check. Density plot of the observed [red] vs. simulated [blue] distribution of beta values. The MSE is approximately 0.241 and supports the idea that the model can be considered to be generating the data.}
\label{fig:prior-predictive-check}
\end{figure}

\begin{figure}[h!]
    \centering
    \includegraphics[width=0.9\textwidth]{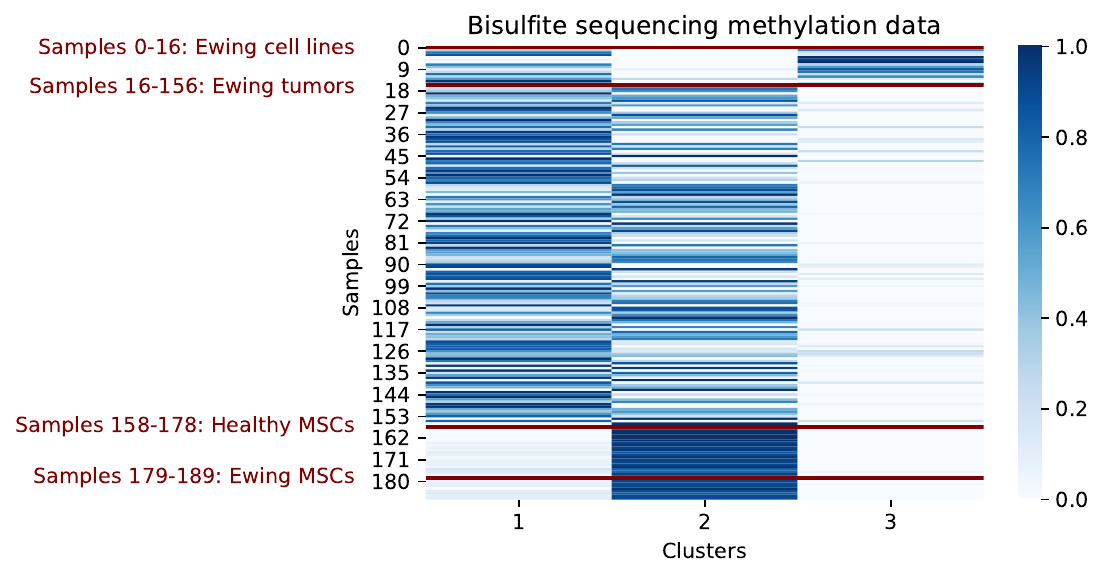}
    \caption{The inferred matrix $\Theta_{I \times C}$. Rows are samples and columns are clusters---each entry represents the strength of association between sample $i$ and cluster $c$.}
    \label{fig:bisulfite-theta}
\end{figure}

\begin{figure}[h!]
    \centering
    \includegraphics[width=.9\textwidth]{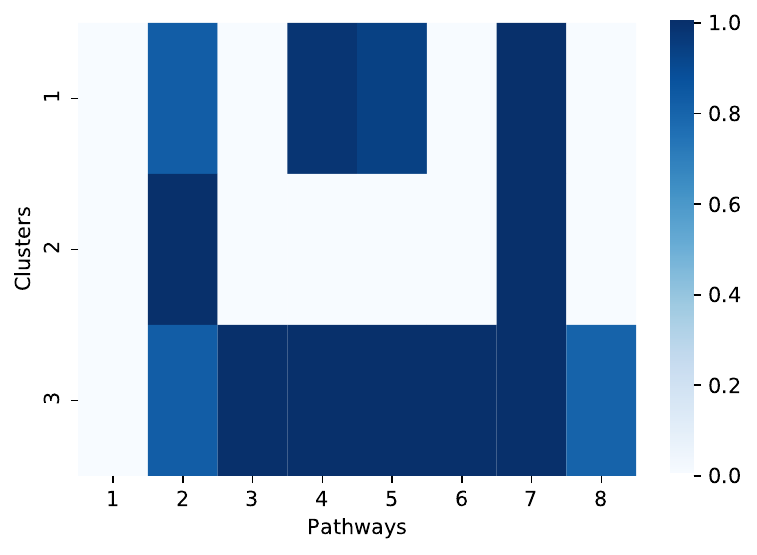}
    \caption{The inferred core matrix $\Pi^{\left( 1 \right)}_{C \times K}$. Each entry represents the strength of association between samples in cluster $c$ and genes in pathway $k$.}
    \label{fig:bisulfite-pi}
\end{figure}

\begin{figure}[h!]
    \centering
    \includegraphics[width=.9\textwidth]{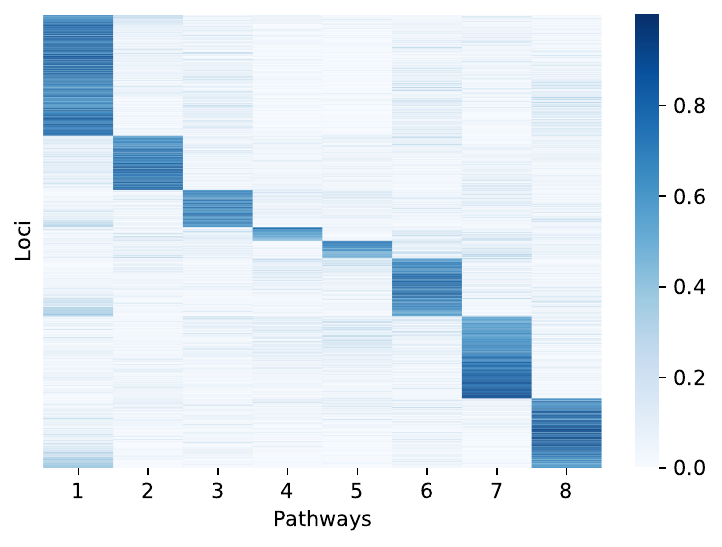}
    \label{fig:bisulfite-phi}
    \caption{The inferred matrix $\Phi_{K \times J}$. Rows are genes and columns are pathways---each entry represents the strength of association between pathway $k$ and gene $j$. Row indices are sorted according to pathway association.}
\end{figure}

\end{document}